\documentclass{article}

\PassOptionsToPackage{numbers, compress}{natbib}
\usepackage[preprint]{neurips_2022}

\usepackage[utf8]{inputenc}

\usepackage{hyperref}
\usepackage{url}
\usepackage{color, soulutf8}
\usepackage[affil-it]{authblk}
\usepackage{amsmath}
\usepackage{comment}
\usepackage{multirow,bigdelim}
\usepackage{lipsum}
\usepackage{array}
\usepackage{wrapfig}

\usepackage{adjustbox}
\usepackage{overpic}
\usepackage{makecell}
\usepackage[normalem]{ulem}
\usepackage{blindtext}
\usepackage{xcolor}
\usepackage{soul}
\usepackage{cleveref}
\usepackage{amsfonts}
\usepackage[linesnumbered, boxed, ruled]{algorithm2e}
\newcolumntype{C}[1]{>{\centering\let\newline\\\arraybackslash\hspace{0pt}}m{#1}}

\newif\ifdraft
\draftfalse

\ifdraft
\definecolor{darkg}{rgb}{0,0.4,0}
\newcommand{\dcc}[1]{{\color{red}[\textbf{DC:} #1]}}
\newcommand{\ahc}[1]{{\color{purple}[\textbf{AH:} #1]}}
\newcommand{\jtc}[1]{{\color{blue}[\textbf{JT:} #1]}}
\newcommand{\kac}[1]{{\color{teal}[\textbf{KA:} #1]}}
\newcommand{\ypc}[1]{{\color{violet}[\textbf{YP} #1]}}
\newcommand{\rmc}[1]{{\color{darkg}[\textbf{RM:} #1]}}

\newcommand{\ron}[1]{{\color{darkg}#1}}

\newcommand{\drop}[1]{}

\else
\newcommand{\ahc}[1]{}
\newcommand{\dcc}[1]{}
\newcommand{\jtc}[1]{}
\newcommand{\kac}[1]{}
\newcommand{\ypc}[1]{}
\newcommand{\rmc}[1]{}

\newcommand{\ron}[1]{{\color{black}#1}}
\fi

\newcommand{\eps}{\varepsilon}
\newcommand{\norm}[1]{\left\Vert #1 \right\Vert_2}

\def\naive{na\"{\i}ve\xspace}

\makeatletter
\DeclareRobustCommand\onedot{\futurelet\@let@token\@onedot}
\def\@onedot{\ifx\@let@token.\else.\null\fi\xspace}

\makeatother

\usepackage[bottom]{footmisc}
\usepackage{arydshln}

\makeatletter
\def\blfootnote{\xdef\@thefnmark{}\@footnotetext}
\makeatother

\newcommand{\attmask}{M}

\newif\ifwatermark
\watermarktrue
\draftfalse
\newcommand*\samethanks[1][\value{footnote}]{\footnotemark[#1]}

\title{Prompt-to-Prompt Image Editing \\ with Cross Attention Control}

\author[1,2]{Amir Hertz\footnote{}~~}
\author[1,2]{Ron Mokady\samethanks ~~}
\author[1]{Jay Tenenbaum}
\author[1]{Kfir Aberman}
\author[1]{Yael Pritch}
\author[1,2]{Daniel Cohen-Or\samethanks ~~}
\affil[1]{ Google Research}
\affil[2]{The Blavatnik School of Computer Science, Tel Aviv University}

\begin{document}

\maketitle

\begin{abstract}

Recent large-scale text-driven synthesis models have attracted much attention thanks to their remarkable capabilities of generating highly diverse images that follow given text prompts.
Such text-based synthesis methods are particularly appealing to humans who are used to verbally describe their intent.
Therefore, it is only natural to extend the text-driven image synthesis to text-driven image editing.
Editing is challenging for these generative models, since an innate property of an editing technique is to preserve most of the original image, while in the text-based models, even a small modification of the text prompt often leads to a completely different outcome. 
State-of-the-art methods mitigate this by requiring the users to provide a spatial mask to localize the edit, hence, ignoring the original structure and content within the masked region.
In this paper, we pursue an intuitive \emph{prompt-to-prompt} editing framework, where the edits are controlled by text only. 
To this end, we analyze a text-conditioned model in depth and observe that the cross-attention layers are the key to controlling the relation between the spatial layout of the image to each word in the prompt. With this observation, we present several applications which monitor the image synthesis by editing the textual prompt only. This includes localized editing by replacing a word, global editing by adding a specification, and even delicately controlling the extent to which a word is reflected in the image. We present our results over diverse images and prompts, demonstrating high-quality synthesis and fidelity to the edited prompts.

\end{abstract}

\footnotetext{Performed this work while working at Google.}

\section{Introduction}

Recently, large-scale language-image (LLI) models, such as Imagen~\cite{saharia2022photorealistic}, DALL·E 2~\cite{ramesh2022hierarchical} and Parti~\cite{yu2022scaling}, have shown phenomenal generative semantic and compositional power, and gained unprecedented attention from the research community and the public eye. 
These LLI models are trained on extremely large language-image datasets and use state-of-the-art image generative models including auto-regressive and diffusion models. 
However, these models do not provide simple editing means, and generally lack control over specific semantic regions of a given image. In particular, even the slightest change in the textual prompt may lead to a completely different output image.

To circumvent this, LLI-based methods \cite{nichol2021glide, avrahami2022blendedlatent, ramesh2022hierarchical}
require the user to explicitly mask a part of the image to be inpainted, and drive the edited image to change in the masked area only, while matching the background of the original image. This approach has provided appealing results, however, the masking procedure is cumbersome, hampering quick and intuitive text-driven editing. Moreover, masking the image content removes important structural information, which is completely ignored in the inpainting process. Therefore, some editing capabilities are out of the inpainting scope, such as modifying the texture of a specific object.

In this paper, we introduce an intuitive and powerful \textit{textual editing} 
method to semantically edit images in pre-trained text-conditioned diffusion models via \textit{Prompt-to-Prompt} manipulations. To do so, we dive deep into the cross-attention layers and explore their semantic strength as a handle to control the generated image. Specifically, we consider the internal \textit{cross-attention maps}, which are high-dimensional tensors that bind pixels and tokens extracted from the prompt text. We find that these maps contain rich semantic relations which critically affect the generated image.

\begin{figure*}[t]
\centering
\ifwatermark
\includegraphics[trim={0 0 0 0},clip,width=\textwidth]{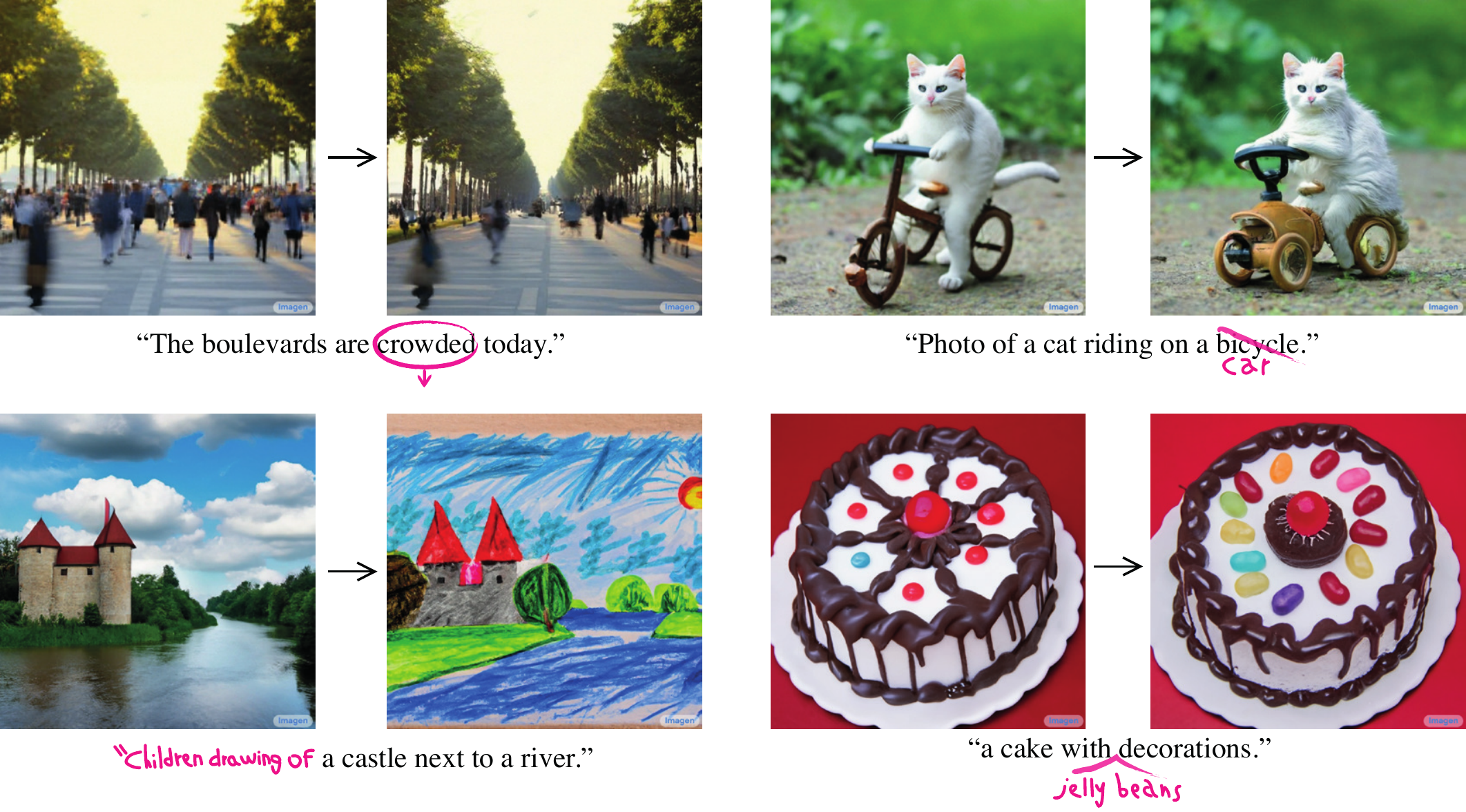}
\else
\includegraphics[trim={0 0 0 0},clip,width=\textwidth]{figures/teaser.pdf}
\fi
\caption{Our method provides variety of \emph{Prompt-to-Prompt} editing capabilities. 
The user can tune the level of influence of an adjective word (top-left), replace items in the image (top-right), specify a style for an image (bottom-left), or make further refinements over the generated image (bottom-right). The manipulations are infiltrated through the cross-attention mechanism of the diffusion model without the need for any specifications over the image pixel space.}
\label{fig:teaser} 
\end{figure*}

Our key idea is that we can edit images by injecting the cross-attention maps during the diffusion process, controlling which pixels attend to which tokens of the prompt text during which diffusion steps. To apply our method to various creative editing applications, we show several methods to control the cross-attention maps through a simple and semantic interface (see \cref{fig:teaser}). The first is to change a single token's value in the prompt (e.g., ``dog" to ``cat"), while fixing the cross-attention maps, to preserve the scene composition. The second is to globally edit an image, e.g., change the style, by adding new words to the prompt and freezing the attention on previous tokens, while allowing new attention to flow to the new tokens. The third is to amplify or attenuate the semantic effect of a word in the generated image.

Our approach constitutes an intuitive image editing interface through editing only the textual prompt, therefore called \textit{Prompt-to-Prompt}. This method enables various editing tasks, which are challenging otherwise, and does not requires model training, fine-tuning, extra data, or optimization. Throughout our analysis, we discover even more control over the generation process, recognizing a trade-off between the fidelity to the edited prompt and the source image. We even demonstrate that our method can be applied to real images by using an existing inversion process. Our experiments and numerous results show that our method enables seamless editing in an intuitive text-based manner over extremely diverse images.

\section{Related work}

\begin{figure*}
\centering
\ifwatermark
\includegraphics[trim={0 0 0 0},clip,width=\textwidth]{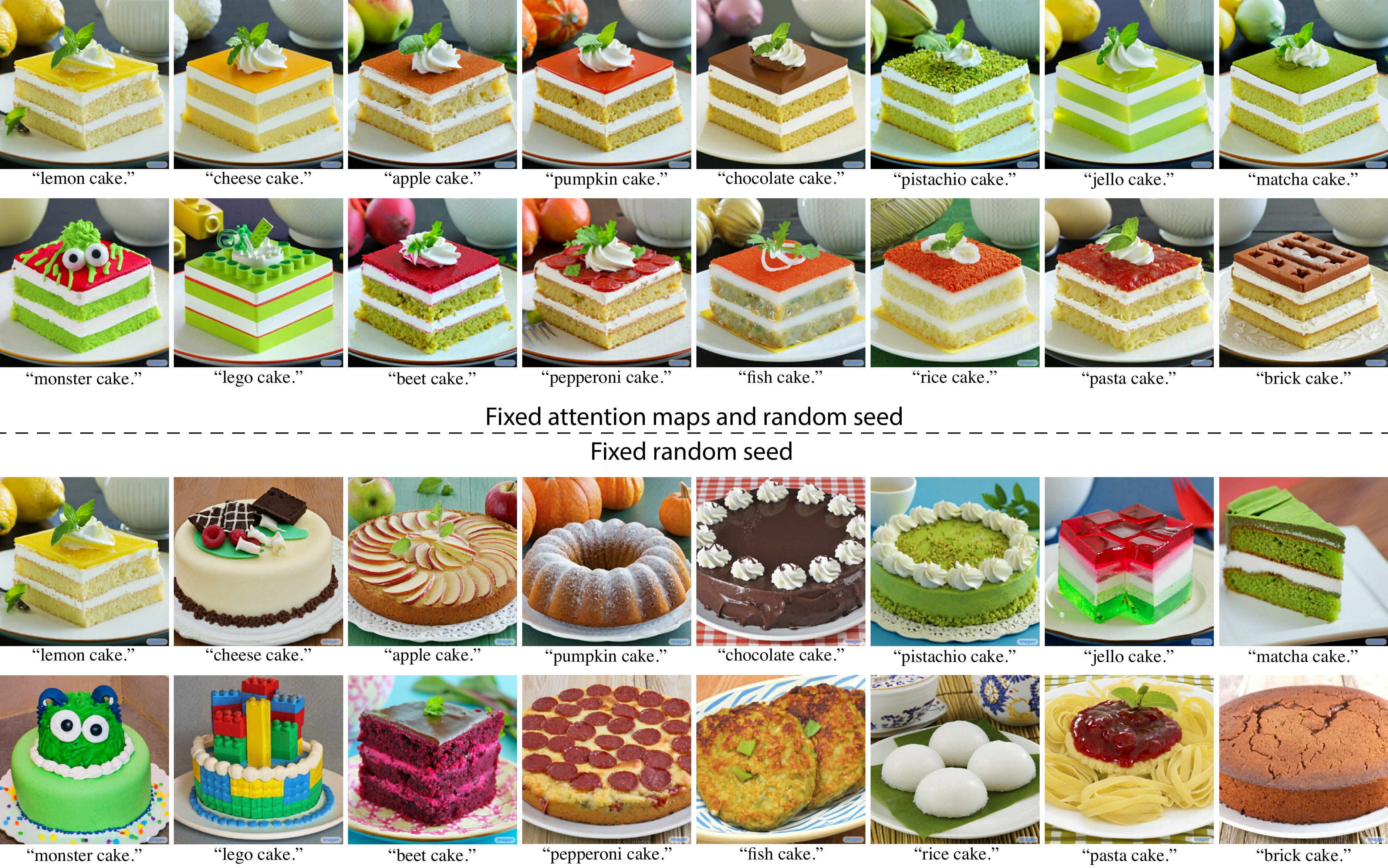}
\else
\includegraphics[trim={0 0 0 0},clip,width=\textwidth]{figures/04_cakes_xl.pdf}
\fi
\caption{Content modification through attention injection. 
We start from an original image generated from the prompt "lemon cake", and modify the text prompt to a variety of other cakes.
On the top rows, we inject the attention weights of the original image during the diffusion process. On the bottom, we only use the same random seeds as the original image, without injecting the attention weights. The latter leads to a completely new structure that is hardly related to the original image.}
\label{fig:cakes} 
\end{figure*}

Image editing is one of the most fundamental tasks in computer graphics, encompassing the process of modifying an input image through the use of an auxiliary input, such as a label, scribble, mask, or reference image. 
A specifically intuitive way to edit an image is through textual prompts provided by the user. 
Recently, text-driven image manipulation has achieved significant progress using GANs
~\cite{goodfellow2014generative,brock2018large,karras2021alias,karras2019style,karras2020analyzing}, which are known for their high-quality generation, in tandem with CLIP~\cite{radford2021learning}, which consists of a semantically rich joint image-text representation, trained over millions of text-image pairs. 
Seminal works \cite{patashnik2021styleclip,gal2021stylegan, xia2021tedigan, abdal2021clip2stylegan} which combined these components were revolutionary, since they did not require extra manual labor, and produced highly realistic manipulations using text only.
Bau et al. \cite{bau2021paint} further demonstrated how to use masks provided by the user, to localize the text-based editing and restrict the change to a specific spatial region. 
However, while GAN-based image editing approaches succeed on highly-curated datasets \cite{mokady2022self}, e.g., human faces, they struggle over large and diverse datasets.

To obtain more expressive generation capabilities, Crowson et al. \cite{crowson2022vqgan} use VQ-GAN \cite{esser2021taming}, trained over diverse data, as a backbone. 
Other works~\cite{avrahami2022blended, kim2022diffusionclip} exploit the recent Diffusion models \cite{ho2020denoising,sohl2015deep,song2019generative,ho2020denoising,song2020denoising,rombach2021highresolution}, which achieve state-of-the-art generation quality over highly diverse datasets, often surpassing GANs~\cite{dhariwal2021diffusion}.
Kim et al.~\cite{kim2022diffusionclip} show how to perform global changes, whereas Avrahami et al.~\cite{avrahami2022blended} successfully perform local manipulations using user-provided masks for guidance.

While most works that require only text (i.e., no masks) are limited to global editing  \cite{crowson2022vqgan, kwon2021clipstyler}, 
Bar-Tal et al.~\cite{bar2022text2live} proposed a text-based localized editing technique without using any mask, showing impressive results.
Yet, their techniques mainly allow changing textures, but not modifying complex structures, such as changing a bicycle to a car.
Moreover, unlike our method, their approach requires training a network for each input.


Numerous works \cite{ding2021cogview,hinz2020semantic,tao2020df, li2019controllable,li2019object,qiao2019learn,qiao2019mirrorgan,ramesh2021zero,zhang2018photographic,crowson2022vqgan, gafni2022make, rombach2021highresolution} significantly advanced the generation of images conditioned on plain text, known as text-to-image synthesis. Several large-scale text-image models have recently emerged, such as Imagen~\cite{saharia2022photorealistic}, DALL-E2~\cite{ramesh2022hierarchical}, and Parti~\cite{yu2022scaling}, demonstrating unprecedented semantic generation. However, these models do not provide control over a generated image, specifically using text guidance only.
Changing a single word in the original prompt associated with the image often leads to a completely different outcome. For instance, adding the adjective ``white'' to ``dog'' often changes the dog's shape.
To overcome this, several works~\cite{nichol2021glide, avrahami2022blendedlatent} assume that the user provides a mask to restrict the area in which the changes are applied. 

Unlike previous works, our method requires textual input only, by using the spatial information from the internal layers of the generative model itself. This offers the user a much more intuitive editing experience of modifying local or global details by merely modifying the text prompt.

\section{Method}
\label{sec:method}

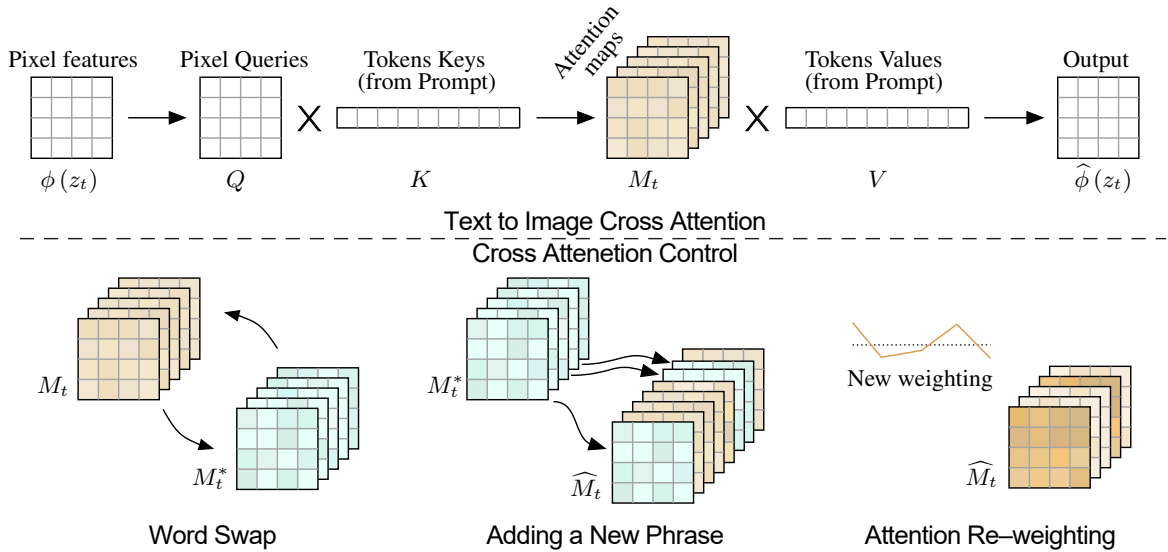
\begin{figure*}[t]
\centering 
\footnotesize{}
\begin{overpic}[width=\textwidth]{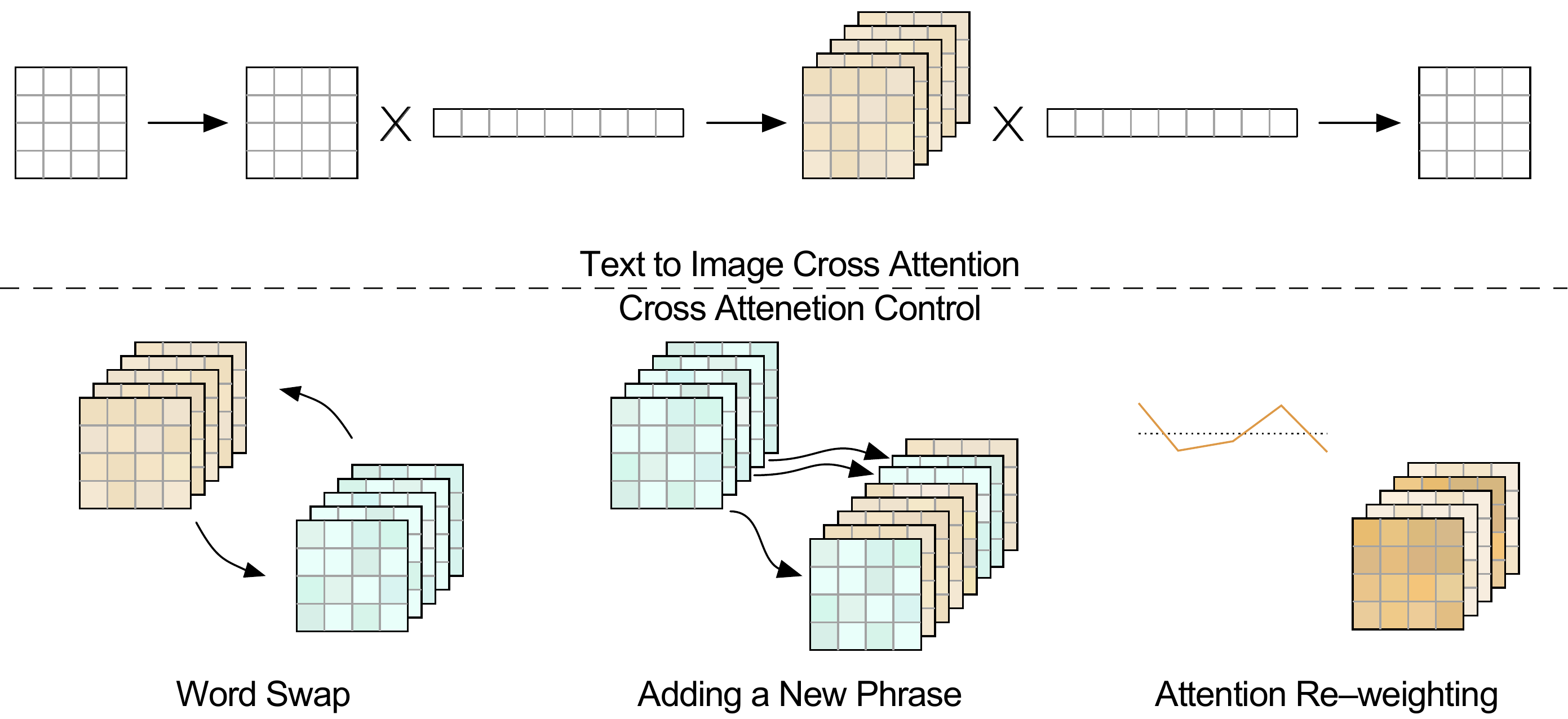} 
\put(1.8,32){$\phi\left(z_t\right) $}
\put(-1,42.5){Pixel features}
\put(14,42.5){Pixel Queries}
\put(30,42.5){Tokens Keys}
\put(29.5,40.5){(from Prompt) }

\put(46.2,41.3){\rotatebox{50}{Attention}}
\put(49.2,41.6){\rotatebox{50}{maps}}
\put(68.5,42.5){Tokens Values}
\put(68.5,40.5){(from Prompt) }
\put(91,42.5){Output}

\put(72.2,14.6){New weighting}

\put(18,32){$Q$}

\put(34,32){$K$}

\put(53,32){$M_{t}$}

\put(74,32){$V$}

\put(92,32){$\widehat{\phi}\left(z_t\right)$}

\put(1.5, 14){$M_t$}

\put(15,6){$M_{t}^{*}$}

\put(35.5, 14){$M_{t}^{*}$}

\put(48,5){$\widehat{M}_{t}$}

\put(82.5,6){$\widehat{M}_{t}$}

\end{overpic}
\caption{Method overview. Top: visual and textual embedding are fused using cross-attention layers that produce spatial attention maps for each textual token. Bottom: we control the spatial layout and geometry of the generated image using the attention maps of a source image. This enables various editing tasks through editing the textual prompt only. When swapping a word in the prompt, we inject the source image maps $M_t$, overriding the target image maps $M^*_t$, to preserve the spatial layout. Where in the case of adding a new phrase, we inject only the maps that correspond to the unchanged part of the prompt. Amplify or attenuate the semantic effect of a word achieved by re-weighting the corresponding attention map.} 
\label{fig:method_diagram} 
\end{figure*}

Let $\mathcal{I}$ be an image which was generated by a text-guided diffusion model \cite{saharia2022photorealistic} using the text prompt $\mathcal{P}$ and a random seed $s$. Our goal is editing the input image guided only by the edited prompt $\mathcal{P}^*$, resulting in an edited image $\mathcal{I}^*$.
For example, consider an image generated from the prompt ``my new bicycle'', and assume that the user wants to edit the color of the bicycle, its material, or even replace it with a scooter while preserving the appearance and structure of the original image.
An intuitive interface for the user is to directly change the text prompt by further describing the appearance of the bikes, or replacing it with another word. As opposed to previous works, we wish to avoid relying on any user-defined mask to assist or signify where the edit should occur. A simple, but an unsuccessful attempt is to fix the internal randomness and regenerate using the edited text prompt. Unfortunately, as \cref{fig:cakes} shows, this results in a completely different image with a different structure and composition.

Our key observation is that the structure and appearances of the generated image depend not only on the random seed, but also on the \emph{interaction} between the pixels to the text embedding through the diffusion process. By modifying the pixel-to-text interaction that occurs in \emph{cross-attention} layers, we provide Prompt-to-Prompt image editing capabilities. More specifically, injecting the cross-attention maps of the input image $\mathcal{I}$ enables us to preserve the original composition and structure. In \cref{sec:crossattentiontext}, we review how cross-attention is used, and in \cref{sec:controlcrossattention} we describe how to exploit the cross-attention for editing. For additional background on diffusion models, please refer to \cref{sec:background}.

\begin{figure*}[t]
\centering
\includegraphics[trim={0 0 0 0},clip,width=\textwidth]{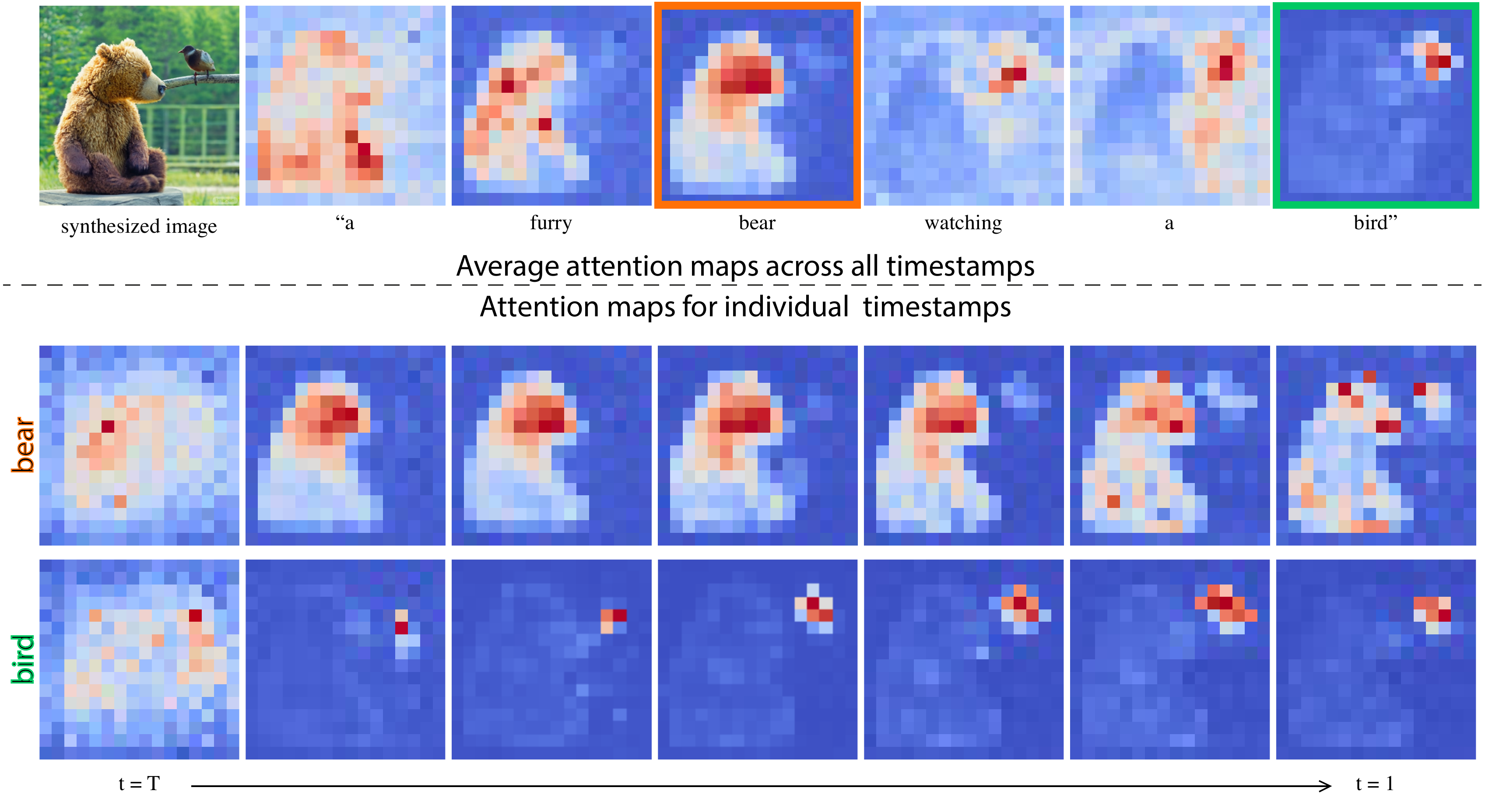}
\caption{Cross-attention maps of a text-conditioned diffusion image generation. The top row displays the average attention masks for each word in the prompt that synthesized the image on the left. The bottom rows display the attention maps from different diffusion steps with respect to the words ``bear'' and ``bird''.}
\label{fig:attention_maps} 
\end{figure*}

\subsection{Cross-attention in text-conditioned Diffusion Models}\label{sec:crossattentiontext}

We use the Imagen~\cite{saharia2022photorealistic} text-guided synthesis model as a backbone. Since the composition and geometry are mostly determined at the $64 \times 64$ resolution, we only adapt the text-to-image diffusion model, using the super-resolution process as is.
Recall that each diffusion step $t$ consists of predicting the noise $\epsilon$ from a noisy image $z_t$ and text embedding $\psi(\mathcal{P})$ using a U-shaped network \cite{ronneberger2015u}. At the final step, this process yields the generated image $\mathcal{I}=z_0$. 
Most importantly, the interaction between the two modalities occurs during the noise prediction, where the embeddings of the visual and textual features are fused using Cross-attention layers that produce spatial attention maps for each textual token.

More formally, as illustrated in \cref{fig:method_diagram}(Top),  the deep spatial features of the noisy image $\phi(z_t)$ are projected to a query matrix $Q = \ell_Q(\phi(z_t))$, and the textual embedding is projected to a key matrix $K = \ell_K(\psi(\mathcal{P}))$ and a value matrix $V = \ell_V(\psi(\mathcal{P}))$, via learned linear projections $\ell_Q, \ell_K, \ell_V$.
The \textit{attention maps} are then 
\begin{equation}
M=\text{Softmax}\left(\frac{QK^T}{\sqrt{d}}\right),
\end{equation}
where the cell $M_{ij}$ defines the weight of the value of the $j$-th token on the pixel $i$, and where $d$ is the latent projection dimension of the keys and queries. Finally, the cross-attention output is defined to be $\widehat{\phi}\left(z_t\right)=MV$, which is then used to update the spatial features $\phi(z_t)$.

Intuitively, the cross-attention output $MV$ is a weighted average of the values $V$ where the weights are the attention maps $M$, which are correlated to the \textit{similarity} between $Q$ and $K$.
In practice, to increase their expressiveness, multi-head attention \cite{NIPS2017_3f5ee243} is used in parallel, and then the results are concatenated and passed through a learned linear layer to get the final output.

Imagen~\cite{saharia2022photorealistic}, similar to GLIDE ~\cite{nichol2021glide}, conditions on the text prompt in the noise prediction of each diffusion step (see \cref{sec:imagentextcondition}) through two types of attention layers: i) cross-attention layers. ii) hybrid attention that acts both as self-attention and cross-attention by simply concatenating the text embedding sequence to the key-value pairs of each self-attention layer. Throughout the rest of the paper, we refer to both of them as cross-attention since our method only intervenes in the cross-attention part of the hybrid attention. That is, only the last channels, which refer to text tokens, are modified in the hybrid attention modules.

\begin{figure*}
\centering
``A photo of a butterfly on...'' \\
\ifwatermark
\includegraphics[trim={0 0 0 0},clip,width=\textwidth]{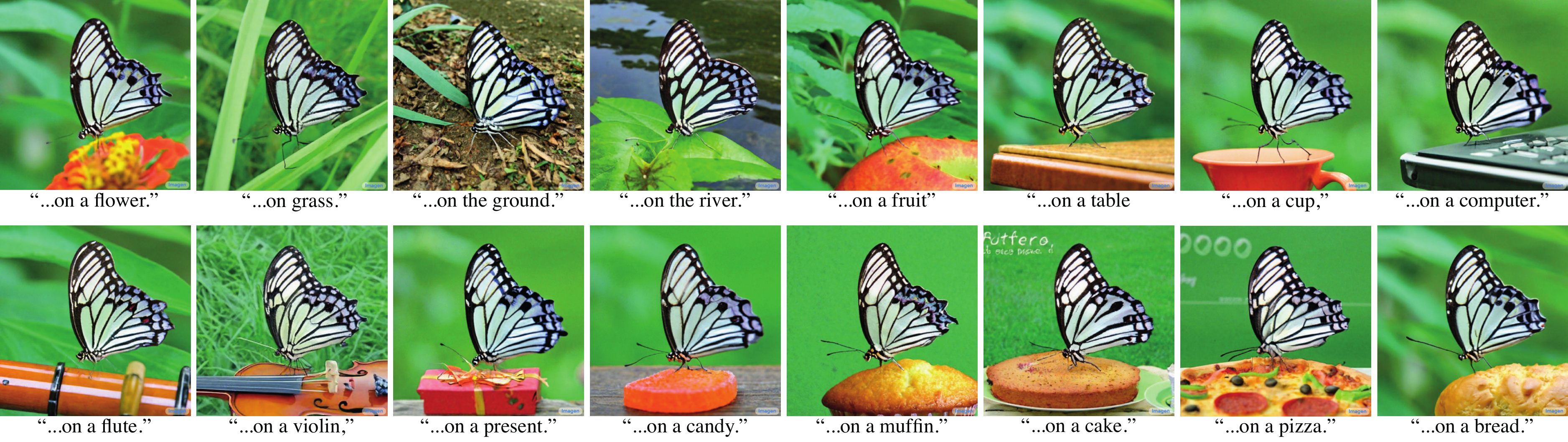} 
\else
\includegraphics[trim={0 0 0 0},clip,width=\textwidth]{figures/04_butterfly.pdf} 
\fi
 
\caption{
Object preservation. By injecting only the attention weights of the word ``butterfly", taken from the top-left image, we can preserve the structure and appearance of a single item while replacing its context. Note how the butterfly sits on top of all objects in a very plausible manner. }
\label{fig:butterfly} 
\end{figure*}

\subsection{Controlling the Cross-attention}\label{sec:controlcrossattention}

We return to our key observation ---  the spatial layout and geometry of the generated image depend on the \emph{cross-attention} maps. This interaction between pixels and text is illustrated in \cref{fig:attention_maps}, where the average attention maps are plotted. As can be seen, pixels are more \textit{attracted} to the words that describe them, e.g., pixels of the bear are correlated with the word ``bear''. Note that averaging is done for visualization purposes, and attention maps are kept separate for each head in our method.
Interestingly, we can see that the structure of the image is already determined in the early steps of the diffusion process.

Since the attention reflects the overall composition, we can inject the attention maps $M$ that were obtained from the generation with the original prompt $\mathcal{P}$, into a second generation with the modified prompt $\mathcal{P}^*$. This allows the synthesis of an edited image $\mathcal{I}^*$ that is not only manipulated according to the edited prompt, but also preserves the structure of the input image $\mathcal{I}$. This example is a specific instance of a broader set of attention-based manipulations leading to different types of intuitive editing. We, therefore, start by proposing a general framework, followed by the details of the specific editing operations.

Let $DM(z_t,\mathcal{P},t,s)$  be the computation of a single step $t$ of the diffusion process, which outputs the noisy image $z_{t-1}$, and the attention map $\attmask_{t}$ (omitted if not used). We denote by $DM(z_t,\mathcal{P},t,s)\{\attmask\gets \widehat{\attmask}\}$ the diffusion step where we override the attention map $\attmask$ with an additional given map $\widehat{\attmask}$, but keep the values $V$ from the supplied prompt. We also denote by $\attmask_{t}^*$ the produced attention map using the edited prompt $\mathcal{P}^*$.
Lastly, we define $Edit(\attmask_t, \attmask_t^*, t)$ to be a general edit function, receiving as input the $t$'th attention maps of the original and edited images during their generation.

\begin{figure*}[t]
\centering
\ifwatermark
\includegraphics[trim={0 0 0 0},clip,width=\textwidth]{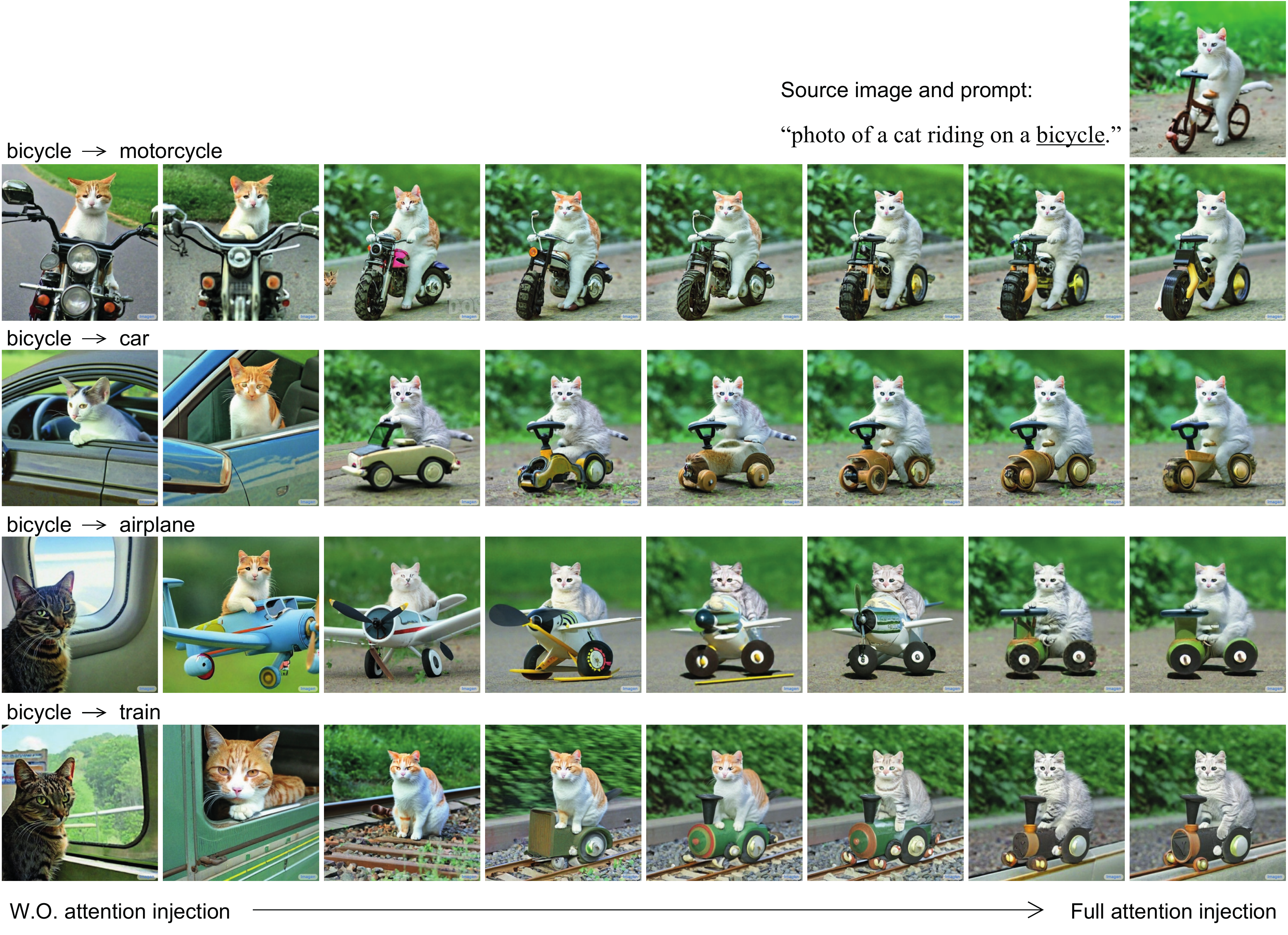}
\else
\includegraphics[trim={0 0 0 0},clip,width=\textwidth]{figures/04_attention_level.pdf}
\fi
\caption{Attention injection through a varied number of diffusion steps. On the top, we show the source image and prompt. In each row, we modify the content of the image by replacing a single word in the text and injecting the cross-attention maps of the source image ranging from 0\% (on the left) to 100\% (on the right) of the diffusion steps. Notice that on one hand, without our method, none of the source image content is guaranteed to be preserved. On the other hand, injecting the cross-attention throughout all the diffusion steps may over-constrain the geometry, resulting in low fidelity to the text prompt, e.g., the car (3rd row) becomes a bicycle with full cross-attention injection.} 
\label{fig:attention_level} 
\end{figure*}

Our general algorithm for controlled image generation consists of performing the iterative diffusion process for both prompts simultaneously, where an attention-based manipulation is applied in each step according to the desired editing task. 
We note that for the method above to work, we must fix the internal randomness. This is due to the nature of diffusion models, where even for the same prompt, two random seeds produce drastically different outputs. Formally, our general algorithm is:

\begin{algorithm}[H]\label{alg:attentionplease}
\SetAlgoLined
\textbf{Input:} A source prompt $\mathcal{P}$, a target prompt $\mathcal{P}^*$, and a random seed $s$.\\
\textbf{Output:} A source image $x_{src}$ and an edited image $x_{dst}$.\\
 $z_{T} \sim N(0,I)$ a unit Gaussian random variable with random seed $s$\; 
 $z_{T}^* \gets z_{T}$; \\
 \For{$t=T,T-1,\ldots,1$}{
    $z_{t-1}, M_{t} \gets DM(z_{t},\mathcal{P},t,s)$\;
    $M_{t}^* \gets DM(z_{t}^*,\mathcal{P}^*,t,s)$\;
    $\widehat{\attmask}_{t} \gets Edit(\attmask_{t}, \attmask_{t}^*, t)$\;
    $z_{t-1}^* \gets DM(z_{t}^*,\mathcal{P}^*,t,s_t)\{\attmask \gets \widehat{\attmask}_{t}\}$\;
 }
 \textbf{Return} $(z_{0},z_{0}^*)$
 \caption{Prompt-to-Prompt image editing}

\end{algorithm}

Notice that we can also define image $\mathcal{I}$, which is generated by prompt $\mathcal{P}$ and random seed $s$, as an additional input. Yet, the algorithm would remain the same. For editing real images, see \cref{sec:app}. 
Also, note that we can skip the forward call in line $7$ by applying the edit function inside the diffusion forward function. Moreover, a diffusion step can be applied on both $z_{t-1}$ and $z_{t}^*$ in the same batch (i.e., in parallel), and so there is only one step overhead with respect to the original inference of the diffusion model.

We now turn to address specific editing operations, filling the missing definition of the $Edit(\attmask_{t}, \attmask_{t}^*, t)$ function.
An overview is presented in \cref{fig:method_diagram}(Bottom).

\begin{figure*}
\centering
``A car on the side of the street.'' \\
\vspace{5pt}
\includegraphics[trim={0 0 0 0},clip,width=\textwidth]{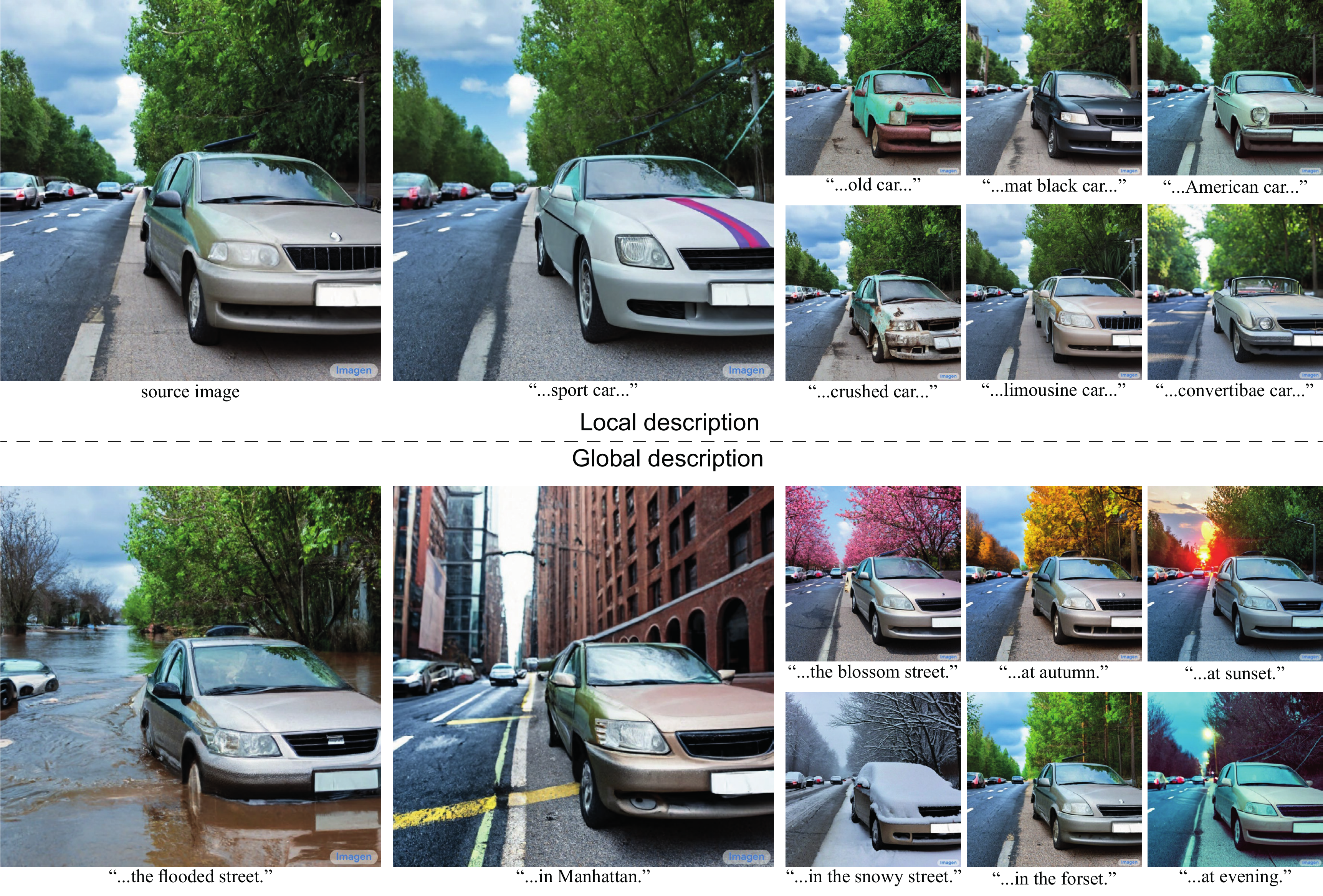}

\caption{Editing by prompt refinement. By extending the description of the initial prompt, we can make local edits to the car (top rows) or global modifications (bottom rows). }
\label{fig:specification} 
\end{figure*}

\paragraph{Word Swap.}
In this case, the user swaps tokens of the original prompt with others, e.g., $\mathcal{P}=$``a big red bicycle'' to $\mathcal{P}^*=$``a big red car''. The main challenge is to preserve the original composition while also addressing the content of the new prompt. To this end, we inject the attention maps of the source image into the generation with the modified prompt. However, the proposed attention injection may over constrain the geometry, especially when a large structural modification, such as ``car'' to ``bicycle'', is involved. We address this by suggesting a softer attention constrain:

$$
Edit(\attmask_{t}, \attmask_{t}^*, t) :=     \begin{cases}
       \attmask_{t}^* &\quad\text{if} \; t < \tau\\
       \attmask_{t} &\quad\text{otherwise.} \\ 
     \end{cases}$$
     where $\tau$ is a timestamp parameter that determines until which step the injection is applied.
    Note that the composition is determined in the early steps of the diffusion process. Therefore, by limiting the number of injection steps, we can guide the composition of the newly generated image while allowing the necessary geometry \textit{freedom} for adapting to the new prompt. An illustration is provided in \cref{sec:app}. Another natural relaxation for our algorithm is to assign a different number of injection timestamps for the different tokens in the prompt. In case the two words are represented using a different number of tokens, the maps can be duplicated/averaged as necessary using an alignment function as described in the next paragraph.

\paragraph{Adding a New Phrase.}
In another setting, the user adds new tokens to the prompt, e.g., $\mathcal{P}=$``a castle next to a river'' to $\mathcal{P}^*=$``children drawing of a castle next to a river''. To preserve the common details, we apply the attention injection only over the common tokens from both prompts. 
Formally, we use an alignment function $A$ that receives a token index from target prompt $\mathcal{P}^*$ and outputs the corresponding token index in $\mathcal{P}$ or \textit{None} if there isn't a match. Then, the editing function is given by:

$$\left(Edit\left(\attmask_{t}, \attmask_{t}^*, t\right)\right)_{i,j}:=     \begin{cases}
       (\attmask_{t}^*)_{i,j} &\quad\text{if} \; A(j) = None\\
       (\attmask_{t})_{i,A(j)} &\quad\text{otherwise.} \\ 
     \end{cases}$$
 Recall that index $i$ corresponds to a pixel value, where $j$ corresponds to a text token.
 Again, we may set a timestamp $\tau$ to control the number of diffusion steps in which the injection is applied. 
This kind of editing enables diverse Prompt-to-Prompt capabilities such as stylization, specification of object attributes, or global manipulations as demonstrated in \cref{sec:app}.

\paragraph{Attention Re--weighting.}
Lastly, the user may wish to strengthen or weakens the extent to which each token is affecting the resulting image. For example, consider the prompt $\mathcal{P}=$ ``a fluffy red ball'', and assume we want to make the ball more or less fluffy. To achieve such manipulation, we scale the attention map of the assigned token $j^*$ with parameter $c\in [-2, 2]$, resulting in a stronger/weaker effect. The rest of the attention maps remain unchanged. That is:
$$\left(Edit\left(\attmask_{t}, \attmask_{t}^*, t\right)\right)_{i,j}:=     \begin{cases}
      c \cdot (\attmask_{t})_{i,j} &\quad\text{if }j=j^*\\
        (\attmask_{t})_{i,j} &\quad\text{otherwise.} \\ 
     \end{cases}$$
As described in \cref{sec:app}, the parameter $c$ allows fine and intuitive control over the induced effect.

\begin{figure*}
\centering
\includegraphics[trim={0 0 0 0},clip,width=\textwidth]{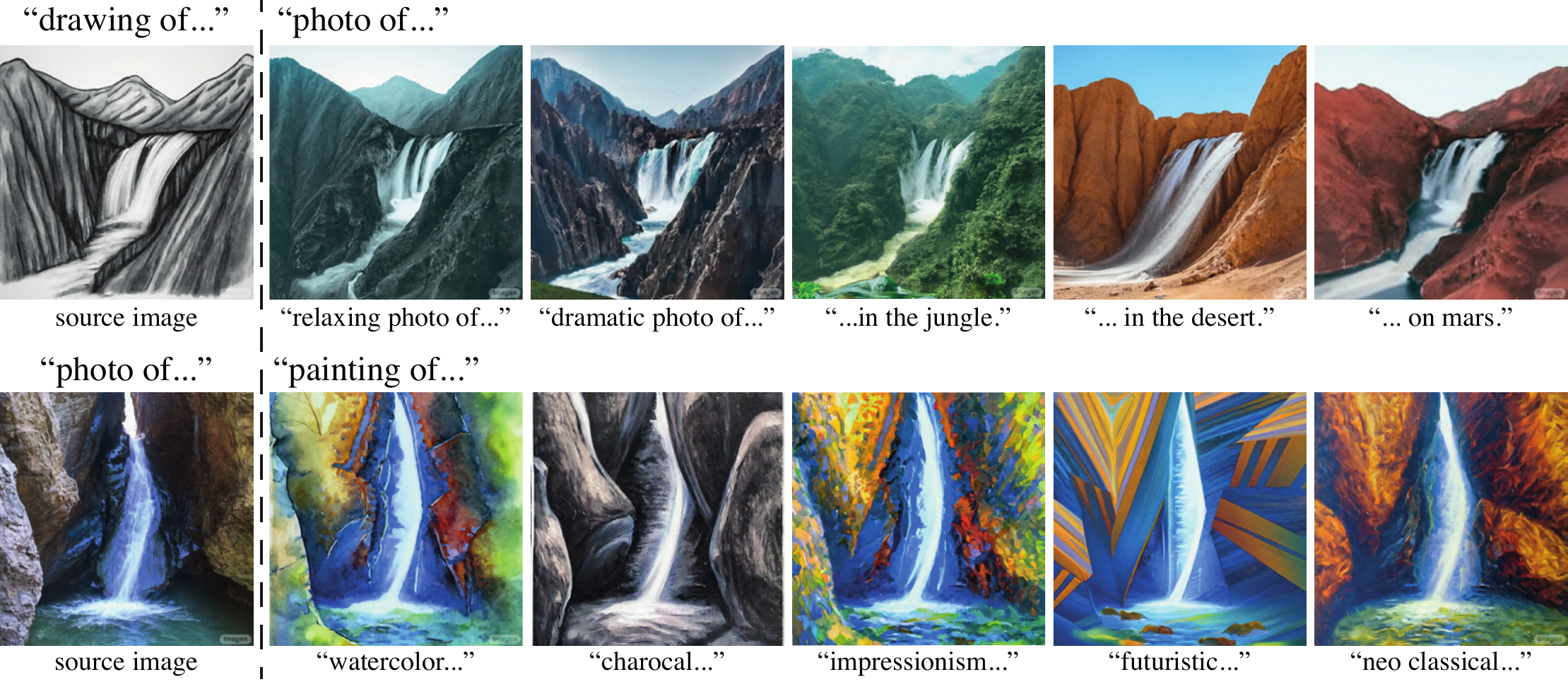}
``A waterfall between the mountains.'' 
\caption{Image stylization. By adding a style description to the prompt while injecting the source attention maps, we can create various images in the new desired styles that preserve the structure of the original image.}
\label{fig:style_transfer} 
\end{figure*}

\section{Applications} 
\label{sec:app}

Our method, described in \cref{sec:method}, enables intuitive text-only editing by controlling the spatial layout corresponding to each word in the user-provided prompt. In this section, we show several applications using this technique.

\textbf{Text-Only Localized Editing.}
We first demonstrate localized editing by modifying the user-provided prompt without requiring any user-provided mask. In \cref{fig:cakes}, we depict an example where we generate an image using the prompt ``lemon cake''. Our method allows us to retain the spatial layout, geometry, and semantics when replacing the word ``lemon'' with ``pumpkin'' (top row). Observe that the background is well-preserved, including the top-left lemons transforming into pumpkins. On the other hand, naively feeding the synthesis model with the prompt ``pumpkin cake'' results in a completely different geometry ($3$rd row), even when using the same random seed in a deterministic setting (i.e., DDIM \cite{song2020denoising}). Our method succeeds even for a challenging prompt such as ``pasta cake.'' ($2$nd row) --- the generated cake consists of pasta layers with tomato sauce on top. Another example is provided in \cref{fig:butterfly} where we do not inject the attention of the entire prompt but only the attention of a specific word -- ``butterfly''. This enables the preservation of the original butterfly while changing the rest of the content. Additional results are provided in the appendix (\cref{fig:supp_word_swapp}).

\begin{figure*}
\centering
\ifwatermark
\includegraphics[trim={0 0 0 0},clip,width=\textwidth]{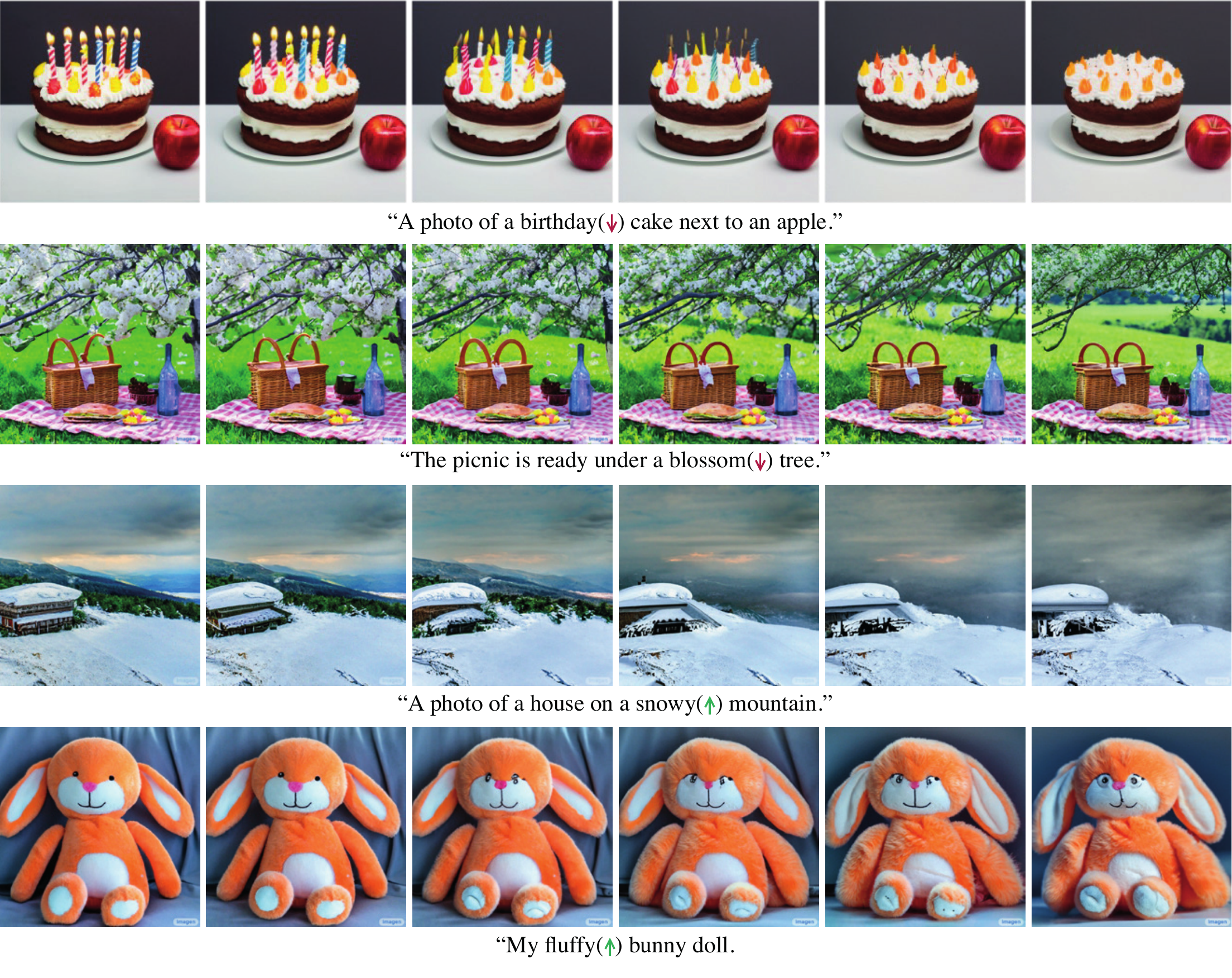}
\else
\includegraphics[trim={0 0 0 0},clip,width=\textwidth]{figures/04_attention_reweight.pdf}
\fi
\caption{Text-based image editing with fader control. By reducing (top rows) or increasing (bottom) the cross-attention of the specified words (marked with an arrow), we can control the extent to which it influences the generated image.}
\label{fig:attention_reweight} 
\end{figure*}

As can be seen in \cref{fig:attention_level}, our method is not confined to modifying only textures, and it can perform structural modifications, e.g., change a ``bicycle'' to a ``car''. To analyze our attention injection, in the left column we show the results without cross-attention injection, where changing a single word leads to an entirely different outcome. From left to right, we then show the resulting generated image by injecting attention to an increasing number of diffusion steps. Note that the more diffusion steps in which we apply cross-attention injection, the higher the fidelity to the original image. 
However, the optimal result is not necessarily achieved by applying the injection throughout all diffusion steps. Therefore, we can provide the user with even better control over the fidelity to the original image by changing the number of injection steps.

Instead of replacing one word with another, the user may wish to add a new specification to the generated image. In this case, we keep the attention maps of the original prompt, while allowing the generator to address the newly added words. For example, see \cref{fig:specification} (top), where we add ``crushed'' to the ``car'', resulting in the generation of additional details over the original image while the background is still preserved. See the appendix (\cref{fig:supp_refine}) for more examples.

\textbf{Global editing.}
Preserving the image composition is not only valuable for localized editing, but also an important aspect of global editing. In this setting, the editing should affect all parts of the image, but still retain the original composition, such as the location and identity of the objects. As shown in \cref{fig:specification} (bottom), we retain the image content while adding ``snow'' or changing the lightning. Additional examples appear in \cref{fig:style_transfer}, including translating a sketch into a photo-realistic image and inducing an artistic style.

\textbf{Fader Control using Attention Re-weighting.}
While controlling the image by editing the prompt is very effective, we find that it still does not allow full control over the generated image. Consider the prompt ``snowy mountain''. A user may want to control the \emph{amount} of snow on the mountain. However, it is quite difficult to describe the desired amount of snow through text. Instead, we suggest a \emph{fader} control \cite{lample2017fader}, where the user controls the magnitude of the effect induced by a specific word, as depicted in \cref{fig:attention_reweight}. As described in \cref{sec:method}, we achieve such control by re-scaling the attention of the specified word. Additional results are in the appendix (\cref{fig:supp_reweight}).

\begin{figure*}[t]
\centering
\ifwatermark
\includegraphics[trim={0 0 0 0},clip,width=\textwidth]{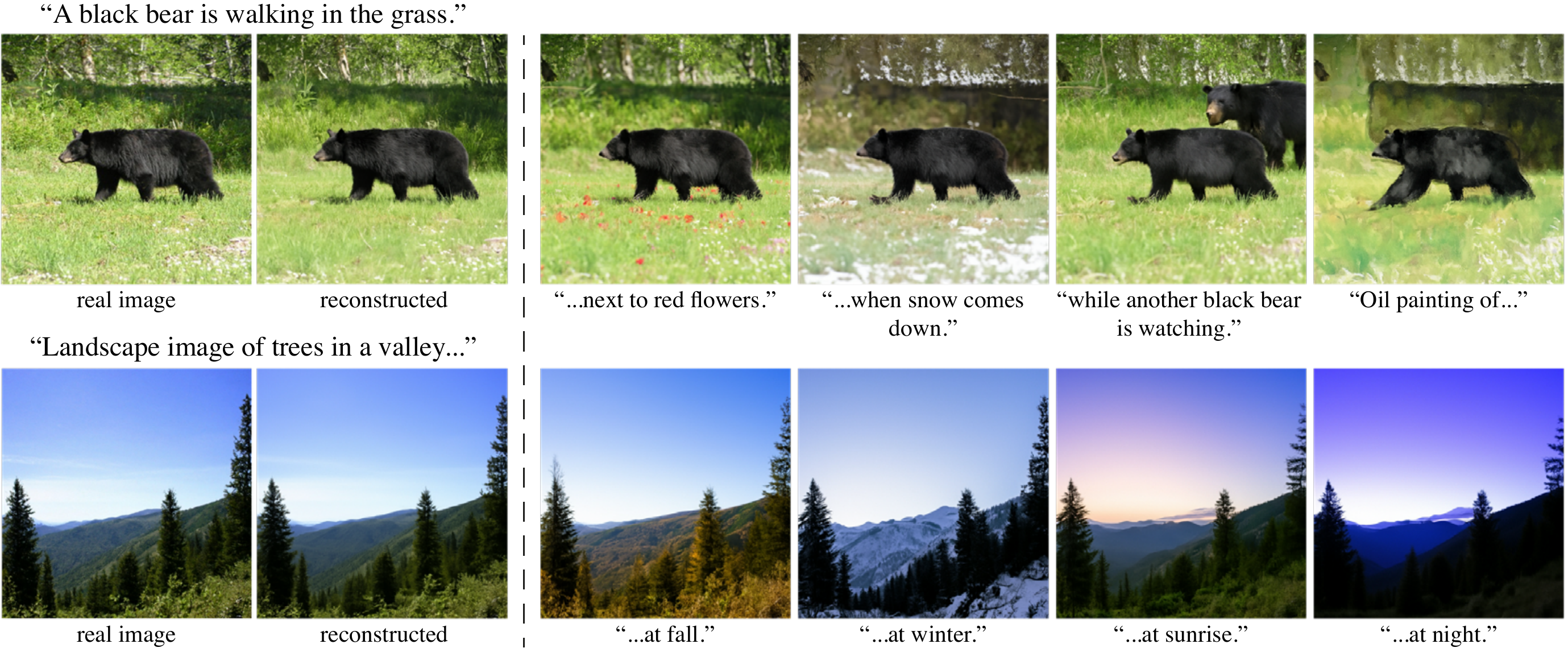}
\else
\includegraphics[trim={0 0 0 0},clip,width=\textwidth]{figures/04_real_images_v2.pdf}
\fi
\caption{Editing of real images. On the left, inversion results using DDIM \cite{song2020denoising} sampling. We reverse the diffusion process initialized on a given real image and text prompt. This results in a latent noise that produces an approximation to the input image when fed to the diffusion process. 
Afterward, on the right, we apply our Prompt-to-Prompt technique to edit the images. }
\label{fig:real_images} 
\end{figure*}

\begin{figure*}[t]
\centering
\includegraphics[trim={0 0 0 0},clip,width=\textwidth]{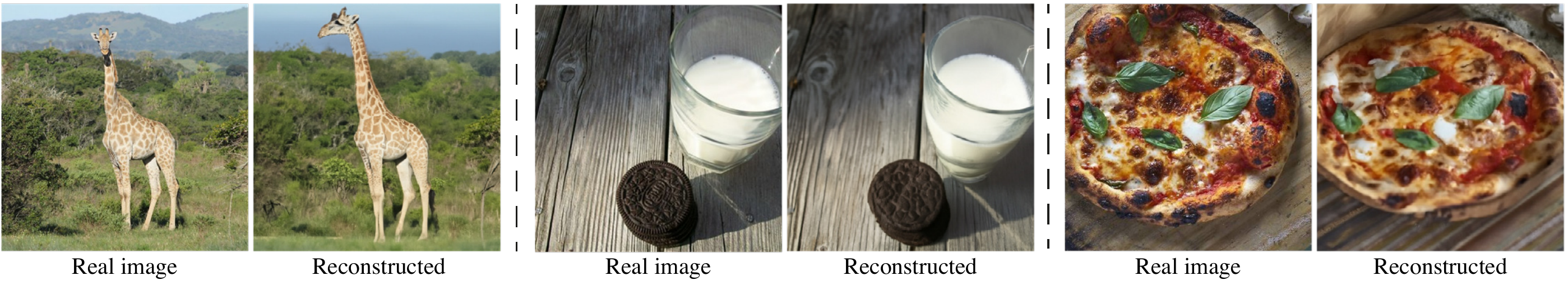}
\caption{Inversion Failure Cases. Current DDIM-based inversion of real images might result in unsatisfied reconstructions. }
\label{fig:limitation} 
\end{figure*}

\textbf{Real Image Editing.}
Editing a real image requires finding an initial noise vector that produces the given input image when fed into the diffusion process. This process, known as \textit{inversion}, has recently drawn considerable attention for GANs, e.g., \cite{zhu2016generative, abdal2019image2stylegan, alaluf2022hyperstyle, roich2021pivotal, zhu2020domain, tov2021designing, Wang2021HighFidelityGI, xia2021gan}, but has not yet been fully addressed for text-guided diffusion models.

In the following, we show preliminary editing results on real images, based on common inversion techniques for diffusion models. First, a rather \naive approach is to add Gaussian noise to the input image, and then perform a predefined number of diffusion steps. Since this approach results in significant distortions,
we adopt an improved inversion approach \cite{dhariwal2021diffusion,song2020denoising}, which is based on the deterministic DDIM model rather than the DDPM model. We perform the diffusion process in the reverse direction, that is $x_0 \longrightarrow x_T$ instead of $x_T \longrightarrow x_0$, where $x_0$ is set to be the given real image.

This inversion process often produces satisfying results, as presented in \cref{fig:real_images}. 
However, the inversion is not sufficiently accurate in many other cases, as in \cref{fig:limitation}.
This is partially due to a distortion-editability tradeoff \cite{tov2021designing}, where we recognize that reducing the classifier-free guidance \cite{ho2021classifier} parameter (i.e., reducing the prompt influence) improves reconstruction but constrains our ability to perform significant manipulations.

\begin{figure*}
\centering
\includegraphics[trim={0 0 0 0},clip,width=\textwidth]{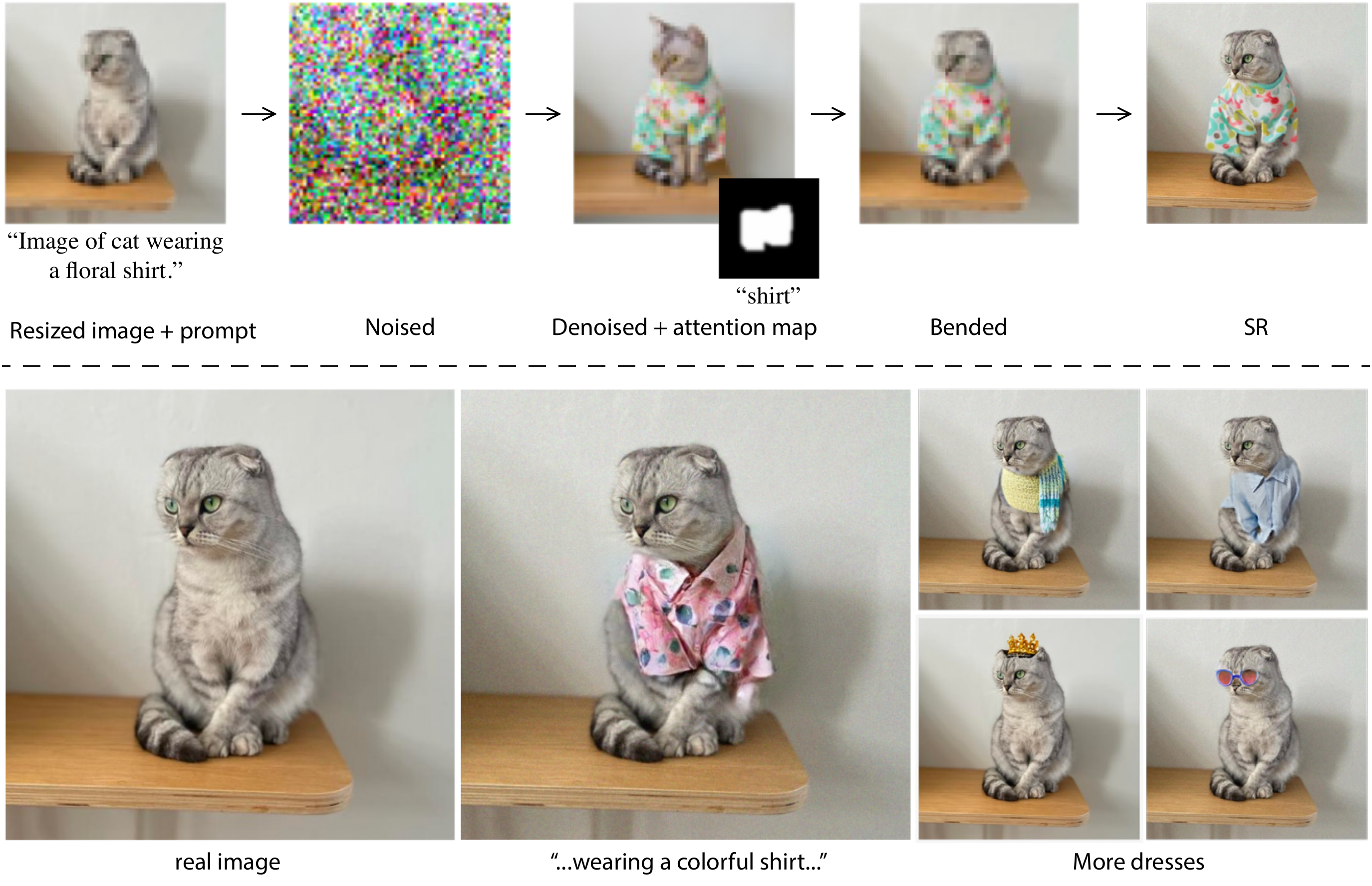} 
\caption{\ron{Mask-based editing. Using the attention maps, we preserve the unedited parts of the image when the inversion distortion is significant. This does not require any user-provided masks, as we extract the spatial information from the model using our method. Note how the cat's identity is retained after the editing process. }}
\label{fig:self_inpainting} 
\end{figure*}

To alleviate this limitation, we propose to restore the unedited regions of the original image using a mask, directly extracted from the attention maps. Note that here the mask is generated with no guidance from the user. As presented in \cref{fig:self_inpainting}, this approach works well even using the \naive DDPM inversion scheme (adding noise followed by denoising). Note that the cat's identity is well-preserved under various editing operations, while the mask is produced only from the prompt itself.

\section{Conclusions}

In this work, we uncovered the powerful capabilities of the cross-attention layers within text-to-image diffusion models.
We showed that these high-dimensional layers have an interpretable representation of spatial maps that play a key role in tying the words in the text prompt to the spatial layout of the synthesized image.
With this observation, we showed how various manipulations of the prompt can directly control attributes in the synthesized image, paving the way to various applications including local and global editing.
This work is a first step towards providing users with simple and intuitive means to \textit{edit} images, leveraging textual semantic power. It enables users to navigate through a semantic, textual, space, which exhibits incremental changes after each step, rather than producing the desired image from scratch after each text manipulation.

While we have demonstrated semantic control by changing only textual prompts, our technique is still subject to a few limitations to be addressed in follow-up work. First, the current inversion process results in a visible distortion over some of the test images. In addition, the inversion requires the user to come up with a suitable prompt.
This could be challenging for complicated compositions. Note that the challenge of inversion for text-guided diffusion models is an orthogonal endeavor to our work, which will be thoroughly studied in the future. Second, the current attention maps are of low resolution, as the cross-attention is placed in the network's bottleneck. This bounds our ability to perform even more precise localized editing. To alleviate this, we suggest incorporating cross-attention also in higher-resolution layers. We leave this for future works since it requires analyzing the training procedure which is out of our current scope. Finally, we recognize that our current method cannot be used to spatially move existing objects across the image and also leave this kind of control for future work.

\section{Acknowledgments}
We thank Noa Glaser, Adi Zicher, Yaron Brodsky and Shlomi Fruchter 
for their valuable inputs that helped improve this work, and to Mohammad Norouzi, Chitwan Saharia and William Chan for providing us with their support and the pretrained models of Imagen \cite{saharia2022photorealistic}.
Special thanks to Yossi Matias for early inspiring discussion on the problem and for motivating and encouraging us to develop technologies along the avenue of intuitive interaction.

\bibliographystyle{plain}
\bibliography{main}

\begin{thebibliography}{10}

\bibitem{abdal2019image2stylegan}
Rameen Abdal, Yipeng Qin, and Peter Wonka.
\newblock Image2stylegan: How to embed images into the stylegan latent space?
\newblock In {\em Proceedings of the IEEE/CVF International Conference on
  Computer Vision}, pages 4432--4441, 2019.

\bibitem{abdal2021clip2stylegan}
Rameen Abdal, Peihao Zhu, John Femiani, Niloy~J Mitra, and Peter Wonka.
\newblock Clip2stylegan: Unsupervised extraction of stylegan edit directions.
\newblock {\em arXiv preprint arXiv:2112.05219}, 2021.

\bibitem{alaluf2022hyperstyle}
Yuval Alaluf, Omer Tov, Ron Mokady, Rinon Gal, and Amit Bermano.
\newblock Hyperstyle: Stylegan inversion with hypernetworks for real image
  editing.
\newblock In {\em Proceedings of the IEEE/CVF Conference on Computer Vision and
  Pattern Recognition}, pages 18511--18521, 2022.

\bibitem{avrahami2022blendedlatent}
Omri Avrahami, Ohad Fried, and Dani Lischinski.
\newblock Blended latent diffusion.
\newblock {\em arXiv preprint arXiv:2206.02779}, 2022.

\bibitem{avrahami2022blended}
Omri Avrahami, Dani Lischinski, and Ohad Fried.
\newblock Blended diffusion for text-driven editing of natural images.
\newblock In {\em Proceedings of the IEEE/CVF Conference on Computer Vision and
  Pattern Recognition}, pages 18208--18218, 2022.

\bibitem{bar2022text2live}
Omer Bar-Tal, Dolev Ofri-Amar, Rafail Fridman, Yoni Kasten, and Tali Dekel.
\newblock Text2live: Text-driven layered image and video editing.
\newblock {\em arXiv preprint arXiv:2204.02491}, 2022.

\bibitem{bau2021paint}
David Bau, Alex Andonian, Audrey Cui, YeonHwan Park, Ali Jahanian, Aude Oliva,
  and Antonio Torralba.
\newblock Paint by word, 2021.

\bibitem{brock2018large}
Andrew Brock, Jeff Donahue, and Karen Simonyan.
\newblock Large scale gan training for high fidelity natural image synthesis.
\newblock {\em arXiv preprint arXiv:1809.11096}, 2018.

\bibitem{crowson2022vqgan}
Katherine Crowson, Stella Biderman, Daniel Kornis, Dashiell Stander, Eric
  Hallahan, Louis Castricato, and Edward Raff.
\newblock Vqgan-clip: Open domain image generation and editing with natural
  language guidance.
\newblock {\em arXiv preprint arXiv:2204.08583}, 2022.

\bibitem{dhariwal2021diffusion}
Prafulla Dhariwal and Alexander Nichol.
\newblock Diffusion models beat gans on image synthesis.
\newblock {\em Advances in Neural Information Processing Systems},
  34:8780--8794, 2021.

\bibitem{ding2021cogview}
Ming Ding, Zhuoyi Yang, Wenyi Hong, Wendi Zheng, Chang Zhou, Da~Yin, Junyang
  Lin, Xu~Zou, Zhou Shao, Hongxia Yang, et~al.
\newblock Cogview: Mastering text-to-image generation via transformers.
\newblock {\em Advances in Neural Information Processing Systems},
  34:19822--19835, 2021.

\bibitem{esser2021taming}
Patrick Esser, Robin Rombach, and Bjorn Ommer.
\newblock Taming transformers for high-resolution image synthesis.
\newblock In {\em Proceedings of the IEEE/CVF conference on computer vision and
  pattern recognition}, pages 12873--12883, 2021.

\bibitem{gafni2022make}
Oran Gafni, Adam Polyak, Oron Ashual, Shelly Sheynin, Devi Parikh, and Yaniv
  Taigman.
\newblock Make-a-scene: Scene-based text-to-image generation with human priors.
\newblock {\em arXiv preprint arXiv:2203.13131}, 2022.

\bibitem{gal2021stylegan}
Rinon Gal, Or~Patashnik, Haggai Maron, Gal Chechik, and Daniel Cohen-Or.
\newblock Stylegan-nada: Clip-guided domain adaptation of image generators.
\newblock {\em arXiv preprint arXiv:2108.00946}, 2021.

\bibitem{goodfellow2014generative}
Ian Goodfellow, Jean Pouget-Abadie, Mehdi Mirza, Bing Xu, David Warde-Farley,
  Sherjil Ozair, Aaron Courville, and Yoshua Bengio.
\newblock Generative adversarial nets.
\newblock {\em Advances in neural information processing systems}, 27, 2014.

\bibitem{hinz2020semantic}
Tobias Hinz, Stefan Heinrich, and Stefan Wermter.
\newblock Semantic object accuracy for generative text-to-image synthesis.
\newblock {\em IEEE transactions on pattern analysis and machine intelligence},
  2020.

\bibitem{ho2020denoising}
Jonathan Ho, Ajay Jain, and Pieter Abbeel.
\newblock Denoising diffusion probabilistic models.
\newblock {\em Advances in Neural Information Processing Systems},
  33:6840--6851, 2020.

\bibitem{ho2021classifier}
Jonathan Ho and Tim Salimans.
\newblock Classifier-free diffusion guidance.
\newblock In {\em NeurIPS 2021 Workshop on Deep Generative Models and
  Downstream Applications}, 2021.

\bibitem{karras2021alias}
Tero Karras, Miika Aittala, Samuli Laine, Erik H{\"a}rk{\"o}nen, Janne
  Hellsten, Jaakko Lehtinen, and Timo Aila.
\newblock Alias-free generative adversarial networks.
\newblock {\em Advances in Neural Information Processing Systems}, 34:852--863,
  2021.

\bibitem{karras2019style}
Tero Karras, Samuli Laine, and Timo Aila.
\newblock A style-based generator architecture for generative adversarial
  networks.
\newblock In {\em Proceedings of the IEEE conference on computer vision and
  pattern recognition}, pages 4401--4410, 2019.

\bibitem{karras2020analyzing}
Tero Karras, Samuli Laine, Miika Aittala, Janne Hellsten, Jaakko Lehtinen, and
  Timo Aila.
\newblock Analyzing and improving the image quality of stylegan.
\newblock In {\em Proceedings of the IEEE/CVF Conference on Computer Vision and
  Pattern Recognition}, pages 8110--8119, 2020.

\bibitem{kim2022diffusionclip}
Gwanghyun Kim, Taesung Kwon, and Jong~Chul Ye.
\newblock Diffusionclip: Text-guided diffusion models for robust image
  manipulation.
\newblock In {\em Proceedings of the IEEE/CVF Conference on Computer Vision and
  Pattern Recognition}, pages 2426--2435, 2022.

\bibitem{kwon2021clipstyler}
Gihyun Kwon and Jong~Chul Ye.
\newblock Clipstyler: Image style transfer with a single text condition.
\newblock {\em arXiv preprint arXiv:2112.00374}, 2021.

\bibitem{lample2017fader}
Guillaume Lample, Neil Zeghidour, Nicolas Usunier, Antoine Bordes, Ludovic
  Denoyer, and Marc'Aurelio Ranzato.
\newblock Fader networks: Manipulating images by sliding attributes.
\newblock {\em Advances in neural information processing systems}, 30, 2017.

\bibitem{li2019controllable}
Bowen Li, Xiaojuan Qi, Thomas Lukasiewicz, and Philip Torr.
\newblock Controllable text-to-image generation.
\newblock {\em Advances in Neural Information Processing Systems}, 32, 2019.

\bibitem{li2019object}
Wenbo Li, Pengchuan Zhang, Lei Zhang, Qiuyuan Huang, Xiaodong He, Siwei Lyu,
  and Jianfeng Gao.
\newblock Object-driven text-to-image synthesis via adversarial training.
\newblock In {\em Proceedings of the IEEE/CVF Conference on Computer Vision and
  Pattern Recognition}, pages 12174--12182, 2019.

\bibitem{mokady2022self}
Ron Mokady, Omer Tov, Michal Yarom, Oran Lang, Inbar Mosseri, Tali Dekel,
  Daniel Cohen-Or, and Michal Irani.
\newblock Self-distilled stylegan: Towards generation from internet photos.
\newblock In {\em Special Interest Group on Computer Graphics and Interactive
  Techniques Conference Proceedings}, pages 1--9, 2022.

\bibitem{nichol2021glide}
Alex Nichol, Prafulla Dhariwal, Aditya Ramesh, Pranav Shyam, Pamela Mishkin,
  Bob McGrew, Ilya Sutskever, and Mark Chen.
\newblock Glide: Towards photorealistic image generation and editing with
  text-guided diffusion models.
\newblock {\em arXiv preprint arXiv:2112.10741}, 2021.

\bibitem{patashnik2021styleclip}
Or~Patashnik, Zongze Wu, Eli Shechtman, Daniel Cohen-Or, and Dani Lischinski.
\newblock Styleclip: Text-driven manipulation of stylegan imagery.
\newblock {\em arXiv preprint arXiv:2103.17249}, 2021.

\bibitem{qiao2019learn}
Tingting Qiao, Jing Zhang, Duanqing Xu, and Dacheng Tao.
\newblock Learn, imagine and create: Text-to-image generation from prior
  knowledge.
\newblock {\em Advances in neural information processing systems}, 32, 2019.

\bibitem{qiao2019mirrorgan}
Tingting Qiao, Jing Zhang, Duanqing Xu, and Dacheng Tao.
\newblock Mirrorgan: Learning text-to-image generation by redescription.
\newblock In {\em Proceedings of the IEEE/CVF Conference on Computer Vision and
  Pattern Recognition}, pages 1505--1514, 2019.

\bibitem{radford2021learning}
Alec Radford, Jong~Wook Kim, Chris Hallacy, Aditya Ramesh, Gabriel Goh,
  Sandhini Agarwal, Girish Sastry, Amanda Askell, Pamela Mishkin, Jack Clark,
  et~al.
\newblock Learning transferable visual models from natural language
  supervision.
\newblock {\em arXiv preprint arXiv:2103.00020}, 2021.

\bibitem{ramesh2022hierarchical}
Aditya Ramesh, Prafulla Dhariwal, Alex Nichol, Casey Chu, and Mark Chen.
\newblock Hierarchical text-conditional image generation with clip latents.
\newblock {\em arXiv preprint arXiv:2204.06125}, 2022.

\bibitem{ramesh2021zero}
Aditya Ramesh, Mikhail Pavlov, Gabriel Goh, Scott Gray, Chelsea Voss, Alec
  Radford, Mark Chen, and Ilya Sutskever.
\newblock Zero-shot text-to-image generation.
\newblock In {\em International Conference on Machine Learning}, pages
  8821--8831. PMLR, 2021.

\bibitem{roich2021pivotal}
Daniel Roich, Ron Mokady, Amit~H. Bermano, and Daniel Cohen-Or.
\newblock Pivotal tuning for latent-based editing of real images.
\newblock {\em ACM Transactions on Graphics (TOG)}, 2022.

\bibitem{rombach2021highresolution}
Robin Rombach, Andreas Blattmann, Dominik Lorenz, Patrick Esser, and Björn
  Ommer.
\newblock High-resolution image synthesis with latent diffusion models, 2021.

\bibitem{ronneberger2015u}
Olaf Ronneberger, Philipp Fischer, and Thomas Brox.
\newblock U-net: Convolutional networks for biomedical image segmentation.
\newblock In {\em International Conference on Medical image computing and
  computer-assisted intervention}, pages 234--241. Springer, 2015.

\bibitem{saharia2022photorealistic}
Chitwan Saharia, William Chan, Saurabh Saxena, Lala Li, Jay Whang, Emily
  Denton, Seyed Kamyar~Seyed Ghasemipour, Burcu~Karagol Ayan, S~Sara Mahdavi,
  Rapha~Gontijo Lopes, Tim Salimans, Tim Salimans, Jonathan Ho, David~J Fleet,
  and Mohammad Norouzi.
\newblock Photorealistic text-to-image diffusion models with deep language
  understanding.
\newblock {\em arXiv preprint arXiv:2205.11487}, 2022.

\bibitem{sohl2015deep}
Jascha Sohl-Dickstein, Eric Weiss, Niru Maheswaranathan, and Surya Ganguli.
\newblock Deep unsupervised learning using nonequilibrium thermodynamics.
\newblock In {\em International Conference on Machine Learning}, pages
  2256--2265. PMLR, 2015.

\bibitem{song2020denoising}
Jiaming Song, Chenlin Meng, and Stefano Ermon.
\newblock Denoising diffusion implicit models.
\newblock In {\em International Conference on Learning Representations}, 2020.

\bibitem{song2019generative}
Yang Song and Stefano Ermon.
\newblock Generative modeling by estimating gradients of the data distribution.
\newblock {\em Advances in Neural Information Processing Systems}, 32, 2019.

\bibitem{tao2020df}
Ming Tao, Hao Tang, Songsong Wu, Nicu Sebe, Xiao-Yuan Jing, Fei Wu, and Bingkun
  Bao.
\newblock Df-gan: Deep fusion generative adversarial networks for text-to-image
  synthesis.
\newblock {\em arXiv preprint arXiv:2008.05865}, 2020.

\bibitem{tov2021designing}
Omer Tov, Yuval Alaluf, Yotam Nitzan, Or~Patashnik, and Daniel Cohen-Or.
\newblock Designing an encoder for stylegan image manipulation.
\newblock {\em arXiv preprint arXiv:2102.02766}, 2021.

\bibitem{NIPS2017_3f5ee243}
Ashish Vaswani, Noam Shazeer, Niki Parmar, Jakob Uszkoreit, Llion Jones,
  Aidan~N Gomez, \L{}ukasz Kaiser, and Illia Polosukhin.
\newblock Attention is all you need.
\newblock In {\em Advances in Neural Information Processing Systems},
  volume~30, 2017.

\bibitem{Wang2021HighFidelityGI}
Tengfei Wang, Yong Zhang, Yanbo Fan, Jue Wang, and Qifeng Chen.
\newblock High-fidelity gan inversion for image attribute editing.
\newblock {\em ArXiv}, abs/2109.06590, 2021.

\bibitem{xia2021tedigan}
Weihao Xia, Yujiu Yang, Jing-Hao Xue, and Baoyuan Wu.
\newblock Tedigan: Text-guided diverse face image generation and manipulation.
\newblock In {\em Proceedings of the IEEE/CVF conference on computer vision and
  pattern recognition}, pages 2256--2265, 2021.

\bibitem{xia2021gan}
Weihao Xia, Yulun Zhang, Yujiu Yang, Jing-Hao Xue, Bolei Zhou, and Ming-Hsuan
  Yang.
\newblock Gan inversion: A survey, 2021.

\bibitem{yu2022scaling}
Jiahui Yu, Yuanzhong Xu, Jing~Yu Koh, Thang Luong, Gunjan Baid, Zirui Wang,
  Vijay Vasudevan, Alexander Ku, Yinfei Yang, Burcu~Karagol Ayan, et~al.
\newblock Scaling autoregressive models for content-rich text-to-image
  generation.
\newblock {\em arXiv preprint arXiv:2206.10789}, 2022.

\bibitem{zhang2018photographic}
Zizhao Zhang, Yuanpu Xie, and Lin Yang.
\newblock Photographic text-to-image synthesis with a hierarchically-nested
  adversarial network.
\newblock In {\em Proceedings of the IEEE conference on computer vision and
  pattern recognition}, pages 6199--6208, 2018.

\bibitem{zhu2020domain}
Jiapeng Zhu, Yujun Shen, Deli Zhao, and Bolei Zhou.
\newblock In-domain gan inversion for real image editing.
\newblock {\em arXiv preprint arXiv:2004.00049}, 2020.

\bibitem{zhu2016generative}
Jun-Yan Zhu, Philipp Kr{\"a}henb{\"u}hl, Eli Shechtman, and Alexei~A Efros.
\newblock Generative visual manipulation on the natural image manifold.
\newblock In {\em European conference on computer vision}, pages 597--613.
  Springer, 2016.

\end{thebibliography}

\appendix
\section{Background}\label{sec:background}

\subsection{Diffusion Models}
Diffusion Denoising Probabilistic Models (DDPM)~\cite{sohl2015deep,ho2020denoising} are generative latent variable models that aim to model a distribution $p_\theta(x_0)$ that approximates the data distribution $q(x_0)$ and easy to sample from. DDPMs model a ``forward process'' in the space of $x_0$ from data to noise.\footnote{This process is called ``forward'' due to its procedure progressing from $x_0$ to $x_T$.} This process is a Markov chain starting from $x_0$, where we gradually add noise to the data to generate the latent variables $x_1,\ldots,x_T\in X$. The sequence of latent variables therefore follows $q(x_1,\ldots,x_t\mid x_0)=\prod_{i=1}^{t}q(x_t\mid x_{t-1})$, where a step in the forward process is defined as a Gaussian transition $q(x_t\mid x_{t-1}):=N(x_t;\sqrt{1-\beta_t}x_{t-1},\beta_t I)$ parameterized by a schedule $\beta_0,\ldots,\beta_T\in (0,1)$. When $T$ is large enough, the last noise vector $x_T$ nearly follows an isotropic Gaussian distribution.

An interesting property of the forward process is that one can express the latent variable $x_t$ directly as the following linear combination of noise and $x_0$ without sampling intermediate latent vectors:
\begin{equation}
  x_t = \sqrt{\alpha_t}x_0+\sqrt{1-\alpha_t}w,~~w\sim N(0,I),\label{eq:xtsamplefromx0}  
\end{equation}
where $\alpha_t:=\prod_{i=1}^{t}(1-\beta_i)$.

In order to sample from the distribution $q(x_0)$, we define the dual ``reverse process'' $p(x_{t-1}\mid x_t)$ from isotropic Gaussian noise $x_T$ to data by sampling the posteriors $q(x_{t-1} \mid x_t)$. Since the intractable reverse process $q(x_{t-1} \mid x_t)$ depends on the unknown data distribution $q(x_0)$, we approximate it with a parameterized Gaussian transition network $p_\theta(x_{t-1}\mid x_t):=N(x_{t-1}\mid \mu_\theta(x_t,t),\Sigma_\theta(x_t,t))$. The $\mu_\theta(x_t,t)$ can be replaced~\cite{ho2020denoising} by predicting the noise $\eps_\theta(x_t,t)$ added to $x_0$ using equation~\ref{eq:xtsamplefromx0}. 

Under this definition, we use Bayes' theorem to approximate
\begin{equation}
\mu_\theta(x_t,t)=\frac{1}{\sqrt{\alpha_t}}\left(x_t-\frac{\beta_t}{\sqrt{1-\alpha_t}}\eps_\theta(x_t,t)\right).
\end{equation}
Once we have a trained $\eps_\theta(x_t,t)$, we can using the following sample method 
\begin{equation}
x_{t-1} = \mu_\theta(x_t,t)+\sigma_t z,~~z\sim N(0,I).
\end{equation}
We can control $\sigma_t$ of each sample stage, and in DDIMs~\cite{song2020denoising} the sampling process can be made deterministic using $\sigma_t=0$ in all the steps. The reverse process can finally be trained by solving the following optimization problem:

$$\min_\theta L(\theta):=\min_\theta E_{x_0\sim q(x_0),w\sim N(0,I),t} \norm{w-\eps_\theta(x_t,t)}^2,$$
teaching the parameters $\theta$ to fit $q(x_0)$ by maximizing a variational lower bound.

\subsection{Cross-attention in Imagen}\label{sec:imagentextcondition}

Imagen \cite{saharia2022photorealistic} consists of three text-conditioned diffusion models: A text-to-image $64\times 64$ model, and two super-resolution models -- $64\times 64 \to 256\times 256$ and $256\times 256 \to 1024\times 1024$.
These predict the noise $\eps_\theta(z_t,c,t)$ via a U-shaped network, for $t$ ranging from $T$ to $1$. Where $z_t$ is the latent vector and $c$ is the text embedding. We highlight the differences between the three models:
\begin{itemize}
    \item \textbf{$64\times 64$} -- starts from a random noise, and uses the U-Net as in \cite{dhariwal2021diffusion}. This model is conditioned on text embeddings via both cross-attention layers at resolutions $[16, 8]$ and hybrid-attention layers at resolutions $[32, 16, 8]$ of the downsampling and upsampling within the U-Net.
    \item \textbf{$64\times 64 \to 256\times 256$} -- conditions on a naively upsampled $64\times 64$ image. An efficient version of a U-Net is used, which includes Hybrid attention layers in the bottleneck (resolution of $32$).
    \item \textbf{$256\times 256 \to 1024\times 1024$} -- conditions on a naively upsampled $256\times 256$ image. An efficient version of a U-Net is used, which only includes cross-attention layers in the bottleneck (resolution of $64$).
\end{itemize}

\begin{figure*}
\centering
\includegraphics[trim={0 0 0 0},clip,width=\textwidth]{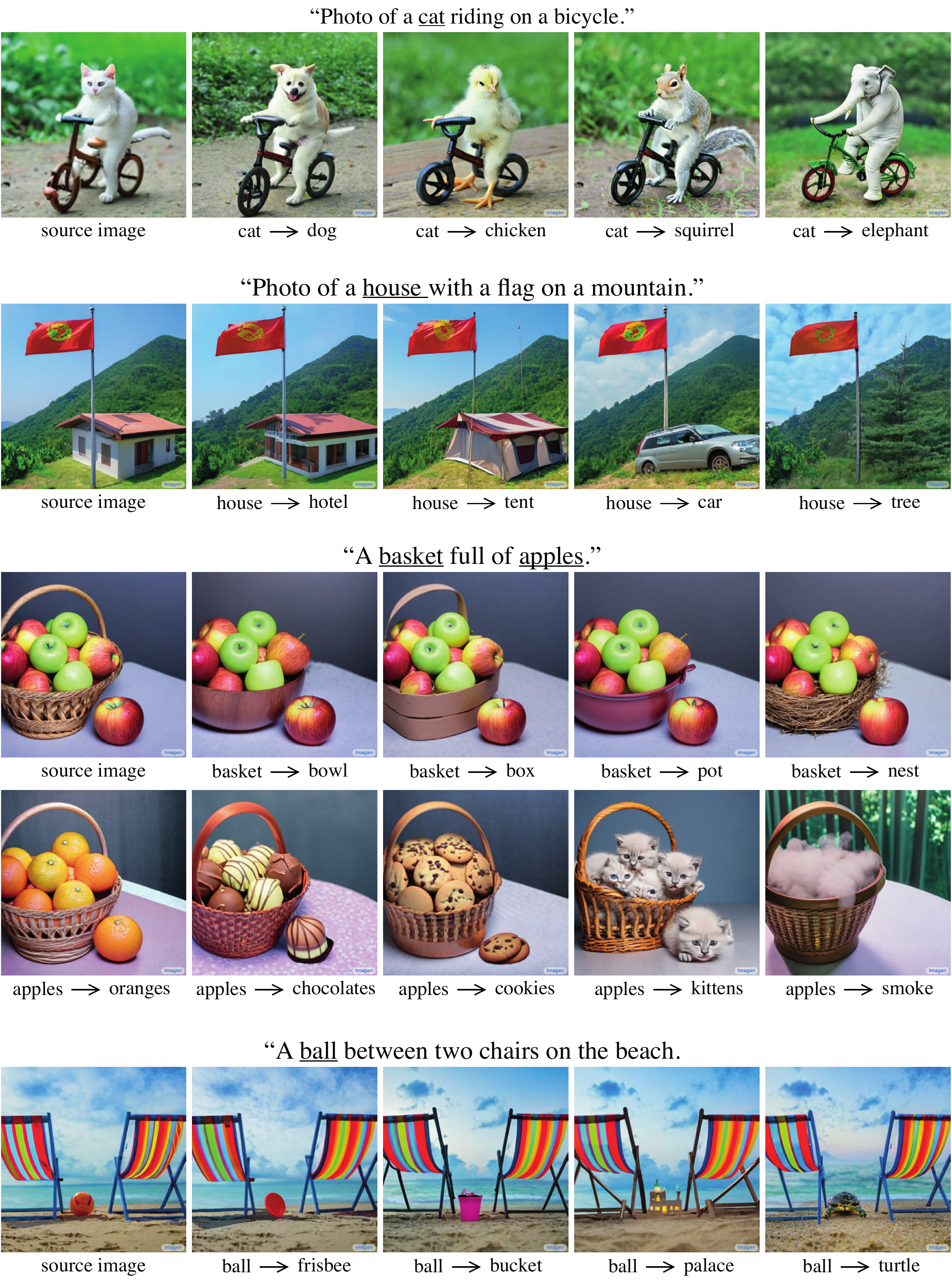}
\caption{Additional results for Prompt-to-Prompt editing by word swap.}
\label{fig:supp_word_swapp} 
\end{figure*}

\section{Additional results} \label{sec:results_supp}

\begin{figure*}
\centering
\ifwatermark
\includegraphics[trim={0 0 0 0},clip,width=\textwidth]{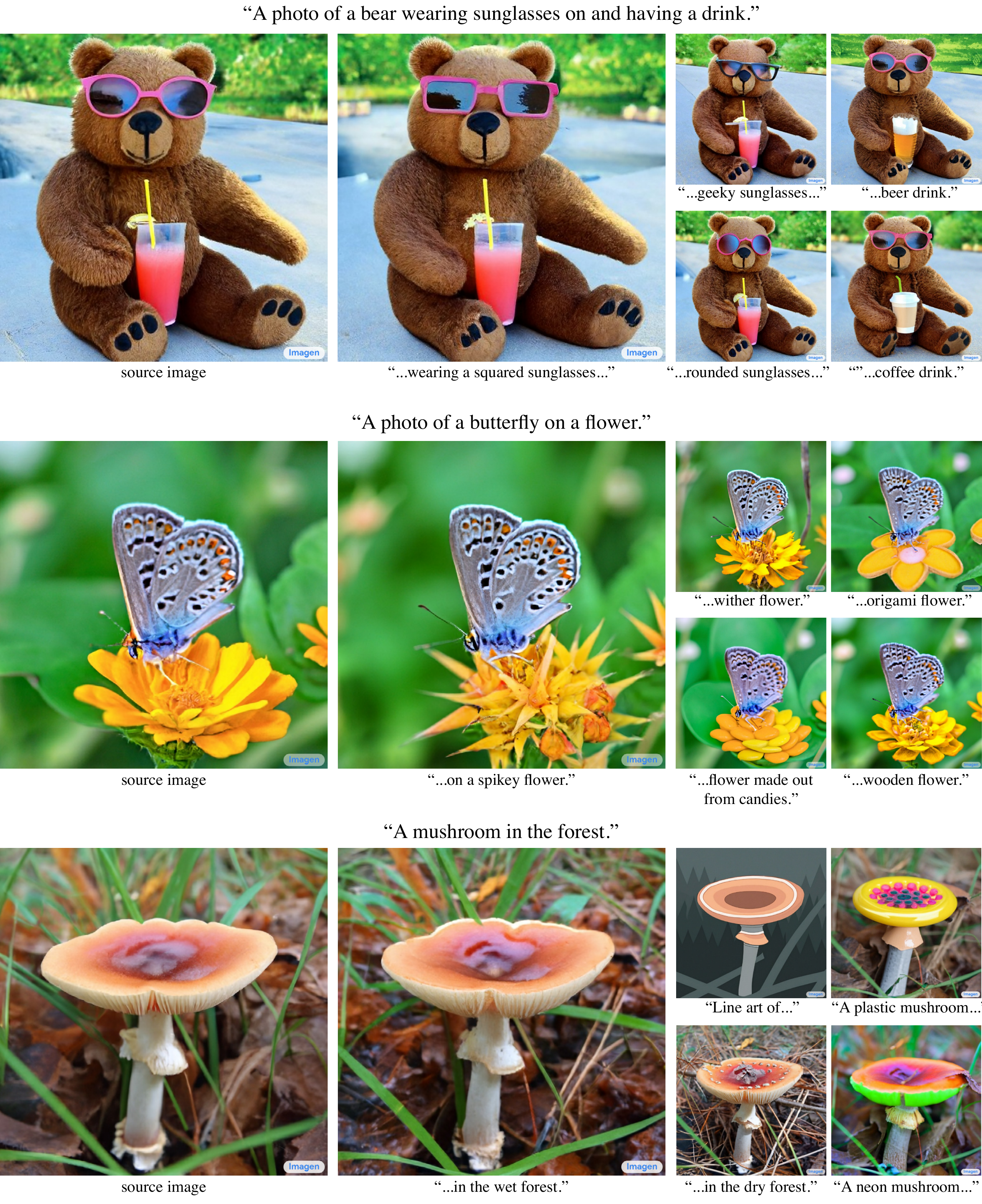}
\else
\includegraphics[trim={0 0 0 0},clip,width=\textwidth]{figures/99_refine_supp.pdf}
\fi
\caption{Additional results for Prompt-to-Prompt editing by adding a specification.}
\label{fig:supp_refine} 
\end{figure*}

\begin{figure*}
\centering
\ifwatermark
\includegraphics[trim={0 0 0 0},clip,width=\textwidth]{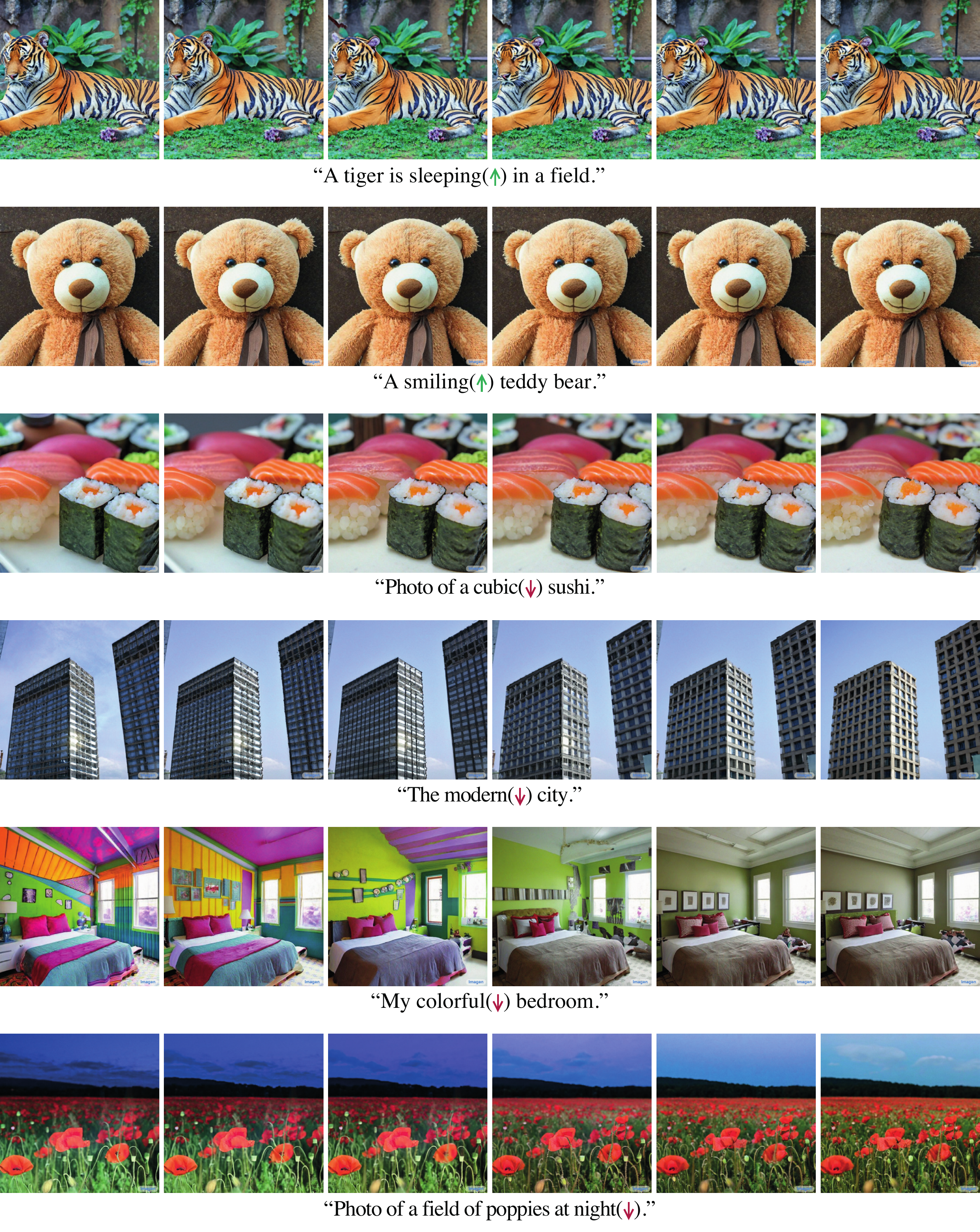}
\else
\includegraphics[trim={0 0 0 0},clip,width=\textwidth]{figures/99_reweight_supp.pdf}
\fi
\caption{Additional results for Prompt-to-Prompt editing by attention re-weighting.}
\label{fig:supp_reweight} 
\end{figure*}

We provide additional examples, demonstrating our method over different editing operations. \cref{fig:supp_word_swapp} show word swap results, \cref{fig:supp_refine} show adding specification to an image, and \cref{fig:supp_reweight} show attention re-weighting.


\end{document}